\documentclass{article}

\usepackage[dvipsnames]{xcolor}
\usepackage{import}

\usepackage{amssymb}
\usepackage{amsthm}

\newtheorem{definition}{Definition}
\newtheorem{examp}{Example}[section]

\usepackage{amsmath}
\usepackage{amsfonts}
\usepackage{makeidx}
\usepackage{graphicx}
\usepackage{float}
\usepackage{hyperref}
\usepackage{stmaryrd}

\newcommand{\mmas}{$\mathcal{MM}$AS}
\newcommand{\+}{~+~}

\newcommand\recc[1]{c?(#1)}
\newcommand\con[1]{$[$cond$]$}

\newcommand\St{St}
\newcommand{\offline}{\textit{offline}}

\newcommand{\Var}{\mbox{\textit{Var}}}
\newcommand{\Data}{\mbox{\textit{Data}}}
\newcommand{\Trans}{\mbox{\textit{Trans}}}
\newcommand{\real}{\mathbb{R}}

\newbox\arriba
\newbox\abajo
\newbox\CaracterInterno
\newbox\CaracterDerecha
\newdimen\anchura
\def\MacrosTranGeneral#1#2#3#4#5#6{%
  \setbox\CaracterInterno=\hbox{\mathsurround=0pt$\mathord#4$}
  \setbox\CaracterDerecha=\hbox{\mathsurround=0pt$\mathord#3$}
  \setbox\arriba=\hbox{$#1#2$}
  \setbox\abajo=\hbox{\mathsurround=0pt%
                      \anchura=\wd\arriba%
                      \advance \anchura by 0.5em%
                      \divide \anchura by \wd\CaracterInterno%
                      \multiply \anchura by \wd\CaracterInterno%
                      \copy\CaracterInterno\kern\SeparacionInternaFlecha
                      \hbox to \anchura{%
                          $\cleaders%
                            \hbox{\kern\SeparacionInternaFlecha\copy\CaracterInterno}
                            \hfill$}%
                      \kern\SeparacionExternaFlecha\copy\CaracterDerecha}
  \mathrel{{\buildrel\vbox{\copy\arriba \kern\SeparacionFlechaArriba} %
    \over{\copy\abajo^{#6}}}_{#5}}
  }
\def\MacrosTranGeneralProp#1#2#3#4#5{\mathchoice%
  {\MacrosTranGeneral{\scriptstyle}{#1}{#2}{#3}{#4}{#5}}
  {\MacrosTranGeneral{\scriptstyle}{#1}{#2}{#3}{#4}{#5}}
  {\MacrosTranGeneral{\scriptscriptstyle}{#1}{#2}{#3}{#4}{#5}}
  {\MacrosTranGeneral{\scriptscriptstyle}{#1}{#2}{#3}{#4}{#5}}}

\def\MacrosTran#1{%
  \def\SeparacionInternaFlecha{-0.3em}
  \def\SeparacionExternaFlecha{-0.5em}
  \def\SeparacionFlechaArriba{-3pt}
  \MacrosTranGeneralProp{#1}{\rightarrow}{-}{}{}}
\def\MacrosNoTran#1{%
  \def\SeparacionInternaFlecha{-0.3em}
  \def\SeparacionExternaFlecha{-0.5em}
  \def\SeparacionFlechaArriba{-3pt}
  \MacrosTranGeneralProp{#1\kern 0.5em}{{\not\rightarrow}}{-}{}{}}

\def\tran#1{\MacrosTran{#1}}

\newcommand{\tranp}[2]{\tran{#1}_{\!\!#2}}
\newcommand{\nat}{\mathbb{N}}
\newcommand{\comen}[1]{}

\usepackage[authoryear,longnamesfirst]{natbib}

\begin{document}

\title{A full process algebraic representation of Ant Colony Optimization\thanks{This paper was published in Information Sciences. The present version is the author's accepted manuscript. Work partially supported by projects
PID2019-108528RB-C22,
and by Comunidad de Madrid as part of the program S2018/TCS-4339 (BLOQUES-CM) co-funded by EIE Funds of the European Union.}}

\author{María García$^{1}$,
        Natalia López$^{2}$,
        and~Ismael Rodríguez$^{2,3}$
\thanks{$^1$Facultad de Ciencias Matemáticas, Universidad Complutense de Madrid, 28040 Madrid, Spain}
\thanks{$^2$
Dpto. Sistemas Inform\'aticos y Computaci\'on.
Facultad de Inform\'atica.
Universidad Complutense de Madrid. 28040 Madrid, Spain.}
\thanks{$^3$Instituto de Tecnolog{\'\i}as del Conocimiento.}
\thanks{E-mail: {\tt mgarci51@ucm.es}, {\tt natalia@sip.ucm.es}, {\tt isrodrig@sip.ucm.es}%
}}

\date{}

\maketitle

\begin{abstract}
We present a process algebra capable of specifying parallelized Ant Colony Optimization algorithms in full detail: PA$^2$CO. After explaining the basis of three different ACO algorithms (Ant System, \textit{MAX\-MIN} Ant System, and Ant Colony System), we formally define PA$^2$CO and use it for representing several types of implementations with different parallel schemes.
In particular fine-grained and coarse-grained specifications, each one taking advantage of parallel executions at different levels of system granularity, are formalized.
\end{abstract}



\maketitle


\section{Introduction}\label{sec1}

Optimization problems appear everywhere in Science, Engineering and Social Sciences. Often these problems are NP-hard, meaning that no algorithm can solve them within reasonable (in particular, polynomial) execution time under standard assumptions (P $\neq$ NP). Thus, sub-optimal heuristic algorithms are typically used instead.
Evolutionary and swarm intelligence algorithms find sub-optimal solutions by iteratively letting a set of simple entities interact with each other according to simple rules, in such a way that the candidate solutions considered by the algorithm tend to improve over time.
For instance, in Genetic Algorithms (GA) (\cite{gol89}) and Particle Swarm Optimization (PSO) (\cite{swarm01,poli07}), these candidate solutions are the entities themselves (chromosomes and particles, respectively), and they iteratively influence each other in such a way that new solutions tend to be more similar to previous good solutions. On the other hand, in Ant Colony Optimization (ACO) (\cite{dorigo03}) and River Formation Dynamics (RFD) (\cite{rrrUC07}), the entities (ants and drops, respectively) collaboratively {\it draw} a solution on a common canvas traversed by all of them together. In particular, in ACO ants traverse a graph while dropping some pheromone trails in the way, and subsequent ants tend to follow paths having higher trails with higher probability. This way, paths in the graphs containing significantly higher pheromone trails are eventually created, and these paths (forming e.g. cycles, trees, etc.) are the solutions for the problems under consideration (e.g. the Traveling Salesman Problem, the Steiner Tree Problem, etc.).

Note that, due to the semi-independent nature of entities in the environment, swarm and evolutionary algorithms in general, and ACO in particular, can be efficiently implemented in parallel (see e.g.~\cite{s19030598,DBLP:journals/tjs/BorisenkoG19,10.1007/s10766-021-00714-1,cmc.2020.012464,mkb2022}). Parallel algorithms are efficient only if each process can perform independent tasks (i.e. tasks not requiring communications with other processes) during a significant proportion of their execution time. ACO fits into this scheme: each process can handle a single ant (or a set of ants) for some time as if the other processes did not exist, and the pheromone trails can be {\it globally} synchronized only once every $n$ iterations of the local processes. With a big $n$, a high level of local independence is achieved, which improves the execution efficiency ---on the cost of using obsolete trail data for longer times, which in turn can harm the convergence of the algorithm to good solutions.

Process algebras are formal languages enabling the systematic analysis of concurrent systems (\cite{milner1992calculus,DBLP:conf/forte/NunezR01,DBLP:conf/voss/LopezN04, fokkink2013introduction}). In them, the parts of the system that are executed sequentially and those executed in parallel (and with which other parts) are explicitly specified, as well as when the latter are required to synchronize. The formal semantics of the language lets us systematically study {\it all} the possible ways in which concurrent events could occur according to the specification. Usually, a {\it labelled transition system}, representing the tree of all possible system executions, can be trivially derived by iteratively applying the operational semantics to the system specification in any possible way. This transition system can be used to systematically search for desirable or undesirable properties (e.g. absence of deadlocks, necessary or possible reachability of some states, termination without starvation, etc.).

Some formal frameworks have been proposed in the literature to enable the definition of evolutionary and swarm intelligent algorithms in general, including some mentions to how to particularize this model to ACO. In some cases, these models do not explicitly represent the concurrent behavior of the system, so the model is the same regardless of whether the execution is sequential or parallel (\cite{DBLP}). In models including an explicit representation of the concurrence (based on process algebras), usually the abstraction level is very high, so the low level details of the ACO behavior (e.g. how the probabilistic choices are made, how the pheromone trails are updated, etc.) are not represented at all (e.g. \cite{eberbach07}).
%
%
Concurrent models based on {\it generalized nets} have been proposed for the representation of ACO~(\cite{fidenovaAtanassov2008,fidenovaAtanassov2009,fidenovaAtanassov2010}), although the algorithm variables, their evolution, and the conditions depending on them are presented in a textual rather than formal form. In these works, the parallelization is introduced so that algorithm variables can evolve concurrently in independent operations. However, since the state of each feature of {\it all} ants is abstracted in the model by a single variable, there is no explicit parallelization of different ants or groups of ants into separated processes, which is the standard approach in \textit{fine-grained} and \textit{coarse-grained} parallel implementations of ACO, respectively.
In addition, some process algebraic models of ACO have been designed to model the behavior of {\it actual} ants in nature, instead of the ACO algorithm itself (e.g.~\cite{tofts1992describing,sbb01, 10.1007/978-3-642-30829-1_11}).

To the best of our knowledge, no formal concurrent model (including process algebras) has been given in the literature providing a full formal definition of the ACO algorithmic scheme and its usual parallel behavior.
The goal of this paper is developing a suitable process algebra language for representing ACO (and possibly also other swarm intelligence methods) and instantiating it to model ACO in {\it full} detail concerning both its functional behavior (i.e. how its actions depend on data previously constructed by the algorithm itself) and its concurrent behavior (i.e. how parallel actions develop both independently and communicating). Three different ACO algorithms are considered: Ant System (\cite{dorigo06}), the 
\textit{MAX-MIN} Ant System (\cite{Sttzle2000MAXMINAS,dorigo03}), and the Ant Colony System (\cite{dorigo96}). Fully detailed specifications of these three algorithms will be defined for our language, and each one will be considered for several levels of parallelism granularity. In each case, the application of the operational semantics to these models will lead {\it exactly} to the behavior of the corresponding algorithms, without any abstractions.
In fact, the process algebra we propose for the modelling of several versions of ACO is designed to precisely include the elements necessary for the representation of the target systems, neither more nor less.%
\footnote{Not surprisingly, the basic language constructions needed for representing the concurrent executions of ACO will be naturally suitable for other swarm intelligence algorithms. This includes the mechanisms combining probabilistic and non-deterministic behaviors, the way information is exchanged between processes, the dependence of probabilities on this information, etc. However, explicitly studying the representation of other swarm intelligence in our language is out of the scope of this paper.}
Thus, its design constitutes an example of system-oriented language construction.
Regarding the models of ACO versions constructed for this language, as far as we know they provide the first fully detailed definition of ACO including a complete representation of both its sequential and its parallel behavior.
In fact, providing fully detailed formal definitions of several parallel implementations of ACO helps in the standardization of these methods: by referring to any of the parallel schemes defined in this paper, a fully unambiguous definition of the assumed kind of parallel scheme is identified, which avoids confusions sometimes induced by informal textual explanations.
Also, we hope this full definition will be the basis to enable a formal systematic analysis of its behavior, as well as serve as an initial step to the full formalization of other swarm and evolutionary algorithms.

It is worth noting that, contrarily to our approach, process algebraic specifications usually adopt a high abstraction of the system being specified. This is useful when they are used to denote e.g., concurrent protocol standards or specifications of the desired behavior of a concurrent system during its first or middle design steps. Moreover, abstract descriptions of complex systems can be analyzed in an automated way more efficiently than the modelled systems themselves.
For instance, if the internal values of some or all variables and their effect on conditions are ignored (abstracted), then the set of possible internal states of the resulting model is dramatically reduced.
In this case, exploring all possible ways of evolution of the system up to some future point through all available choices is much cheaper ---on the cost that the model will non-deterministically represent some choices which were deterministic before the variables these choices depend on were ignored.
Our fully defined specification provides a departure point to easily enable the abstraction kind one would need in each case. By ignoring the variables controlling some aspect of the system behavior or another (e.g. those controlling how many ants must finish before each update, those keeping track of previously traversed nodes, or even those representing the pheromone trail values), we can obtain different abstractions of the original system ---all of them easier to systematically study for all possible executions up to some point than the original full specification, but each one missing behavior details in a different aspect. Making  a detailed model more abstract is, obviously, much easier than adding details to an abstract model, so our specification is a suitable starting model to reach any tailored abstraction one would need in each case.



The rest of the paper is organized as follows. In Section~\ref{sec:ACO} we study the operation and motivation of ACO algorithms, and explain the three variants we will model later (Ant
System, \textit{MAX-MIN} Ant System and Ant Colony System) using the Traveling Salesman Problem as an example of application. Also in this section, we study the main strategies that can be found in literature for parallelizing ACO algorithms: the fine-grained parallelization and the coarse-grained parallelization. In Section~\ref{sec:PA} we define the Process Algebra for ACO, PA$^2$CO, the algebra whose specific goal is to be used   for representing ACO algorithms in full detail.
After that, we formalize the mentioned ACO algorithms in their \textit{fine-grained} and \textit{coarse-grained} parallelism versions in Section~\ref{sec:fine-grained} and Section~\ref{sec:coarse-grained}, respectively. In Section~\ref{sec:discussion} we present a comparison of all the specifications provided in the paper. Finally, we present the main conclusions of this article in Section~\ref{sec:concl}, as well as the proposed future work.




	\mbox{}
	\section{ACO Algorithms }
	\label{sec:ACO}
	
	Swarm intelligence is a consolidated heuristic approach to tackle hard optimization problems, and it is mainly inspired by animal flock interactions (e.g. birds or insects). In particular, the behavior of ants has inspired a large number of algorithms that are known as \textit{Ant Colony Optimization} (\cite{dorigo96}). All these algorithms are based on the same principle: how ants communicate in their search for food.
	
	Ants have a very efficient way of looking for food. When the search starts, ants move randomly. The moment one of them finds food, it returns to the anthill, leaving in its way a trail of pheromones. Since ants are attracted by these trails, this increases the probability that subsequent ants will follow this path, which in turn increases the amount of pheromone in the path. As shorter paths are traversed more often per unit of time than longer ones, and pheromone trails  disappear  over time, the shortest paths to the food will eventually be the ones with the strongest pheromone trail. In this way, all ants will end up following the same path, which is the shortest one between the anthill and the food.
	
	ACO algorithms rely on this process to optimize different types of problems. In this section we will study three of these algorithms: the Ant System (\cite{dorigo06}), the 
    \textit{MAX-MIN} Ant System (\cite{Sttzle2000MAXMINAS,dorigo03}), and the Ant Colony System (\cite{dorigo96}). We will explain them with a typical example of application: the Traveling Salesman Problem (TSP). This problem consists in visiting $n$ cities, each of them exactly once, while minimizing the addition of costs due to moving between cities. More formally, it consists in finding the minimum-cost Hamiltonian path in a complete weighted graph composed of  $n$ vertices ---hence, vertices are the cities, edges are their connections, and their associated weights are the corresponding travelling costs. In the remainder of this text, we will use the variables of Table \ref{tab:setsTSP} to represent these components and the amount of pheromone present in each edge at each moment.
	
	\begin{table}[h]
		
	    \hrulefill
	
	    \centering
	    {\footnotesize
		\begin{tabular}{l}
		    \begin{tabular}{l}
			$v_i =$ vertex corresponding to the city $i$\\
			$e_{i,j} =$ edge joining the vertices $v_i$ and $v_j$\\
			$d_{i,j} =$ weight of the edge $e_{i,j}$\\
			$\tau_{i,j}(t) =$ quantity of pheromone on $e_{i,j}$ at time $t$\\
		    \end{tabular}
		    $\forall$ $i,j\in \{1,\cdots,n\}$ and $i \neq j$
		\end{tabular}\\
		}
		
		\hrulefill
	    \caption{Set of variables of TSP.}
	    \label{tab:setsTSP}
	    \end{table}

	\begin{subsection}{Ant System}\label{aco-as}\label{subsec:AS}
	The first algorithm inspired by the mentioned behavior of ants was proposed in the 90s by Marco Dorigo in his doctoral thesis and it is known as \textit{Ant System} (AS) (\cite{dorigo06}). As we said above, we use the TSP to explain the mechanics of this algorithm.
	
	In the AS algorithm there are several ants traversing the graph and generating Hamiltonian paths. Unlike natural ants, these ants have memory and live in an environment where time is discrete. Memory helps them remember which vertices they have already visited (so as not to visit them again) and, as time is discrete, the decisions made at the moment $t$ are executed at the moment $t + 1$. Each ant  represents a solution (partial if it has not yet passed through all the vertices, and total if it has made a complete lap).
	
	The algorithm distributes a number of ants (in our case, $m$ ants, being $m$ a natural number fixed previously) among the vertices of the graph so that each of them decides which vertex is visited next based on some probabilities. These probabilities depend on both the intensity of the trail and the cost of each edge. Thus, there is a certain balance between choosing the edges most used by other ants, which yields a self-induced feedback process, and the edges with lowest cost, which implicitly poses a trivial greedy strategy by itself.
    In addition, as each ant remembers the vertices that it has already visited, the probability of choosing edges leading to a repeated vertex can be set to zero, as it is required by the restrictions of the problem. In particular, the formula for the probability that the k-th ant chooses the edge $e_{i,j}$ at the moment $t$ proposed in the AS is:
	\begin{equation}\label{prob-as}
		p_{i,j}^k(t) = \left\{
		\begin{array}{lll}
			\frac{[\tau_{i,j}(t)]^\alpha \cdot [\eta_{i,j}]^\beta}{\Sigma_{r \in \textit{allowed}_k} [\tau_{i,r}(t)]^\alpha \cdot [\eta_{i,r}]^\beta} & $if~$ v_j \notin path_k\\[0.5em]
						0                  & $otherwise$\\
		\end{array}
		\right.
	\end{equation}
	where:
	\begin{itemize}
		\item $\eta_{i,j} =  \frac{1}{d_{i,j}}$ is the visibility of the edge $e_{i,j}  $,
		\item $\alpha, \beta$ are the parameters to control the relative importance between trail and visibility, and
		\item $\textit{path}_k$ is the list of vertices that have already been visited by the k-th ant.
	\end{itemize}
	\indent Once the ant has chosen which vertex to move to at moment $t$, it executes the movement at instant $t + 1$. When it is already there, it chooses probabilistically again. At every moment of time each ant decides and moves, so at moment $t+n$ all the ants will have completed one lap (or Hamiltonian path). At this moment, the AS algorithm updates the trail of each edge $e_{i,j}$ according to the following formula:
	\begin{equation}\label{rastros-as}
		\tau_{i,j}(t+n) = (1 - \rho) \cdot \tau_{i,j}(t) + \Sigma_{k = 1}^m \Delta\tau_{i,j}^k
	\end{equation}
	where:
	\begin{itemize}
		\item $\rho$ is the pheromone decay coefficient in the time interval $(t,t+n)$,
		\item $\Delta\tau_{i,j}^k = \left\{ \begin{array}{lll}
			\frac{Q}{L_k} & $if the k-th ant uses the edge $e_{i,j}$ on its lap$\\[0.5em]
				0          & $otherwise$\\
		\end{array}
		\right.$
		\item $Q$ is a constant, and
        \item $L_k$ is the total length of the lap of the k-th ant.
	\end{itemize}
	\indent The values $\tau_{i,j}(0)$ must be initialized at the beginning of the algorithm, and it is common to give them very low values. Intuitively, the formula expresses that the trail of the edge $e_{i,j}$ after a complete lap of the ants will be the amount of pheromone present at the moment $t$ that has not been evaporated, plus the pheromones that have been left by all the ants that have passed through that edge in their path.\\ \indent Finally, the algorithm will end, either when the maximum number of iterations is reached, or when all the ants perform the same path.

	\begin{subsection}{\textit{MAX\-MIN} Ant System}\label{subsec:maxminAS}
	
	The difference between the \textit{MAX\-MIN} Ant System, abbreviated $\mathcal{MM}$AS (\cite{Sttzle2000MAXMINAS,dorigo03}), and the AS is how trails are updated. In the AS the amount of pheromone on each edge is updated by all the ants that have visited it in their lap, whereas in the $\mathcal{MM}$AS the only ant that updates the trails is the one whose path has the minimum cost, which introduces a kind of {\it elitism}. In addition, in this version the values of the trails are bounded to avoid stagnation of the search algorithm. Thus, the update of the trail is as follows:
	\begin{equation}\label{rastros-mmas}
		\tau_{i,j}(t+n) = \bigg[ (1 - \rho) \cdot \tau_{i,j}(t) + \Delta \tau_{i,j}^{\textit{best}}  \bigg]_{ \tau_{\textit{min}} }^{ \tau_{\textit{max}} }
	\end{equation}
	where:
	\begin{itemize}
		\item $\Delta \tau_{i,j}^{\textit{best}} = \left\{
		\begin{array}{lll}
			\frac{1}{L_{\textit{best}}} & {\rm if\ }e_{i,j} \in \textit{edges}_{best} \\[0.5em]
			0                  & $otherwise$\\
		\end{array}
		\right.\\[.3em]$ being $\textit{edges}_{best}$ the list of edges chosen by the ant with the best path,
		\item $L_{best}$ is the length of the path with minimum cost,
		\item $\tau_{\textit{min}}, \tau_{\textit{max}}$ is the minimum and maximum trail quantity values,
	\end{itemize}
	and the operator $\big[ ~~ \big]_b^a$ is defined as:
	\begin{itemize}
		\item $\big[ x \big]_b^a = \left\{
		\begin{array}{lll}
			a & $if $ x > a\\
			x & $if $ b \leq x \leq a\\
			b & $if $ x < b\\
		\end{array}
		\right.$
	\end{itemize}

\end{subsection}

\begin{subsection}{Ant Colony System}\label{aco-acs}\label{subsec:ACS}
	
	Regarding the Ant Colony System, ACS (\cite{dorigo96}), its greatest contribution is the introduction of a local update of the amount of pheromone in addition to the one that is carried out in the AS after each complete lap, which we will now refer to as an \textit{offline} update. Thus, each ant probabilistically chooses an edge $e_{i,j}$, and immediately after that, the trail $\tau_{i,j}$ is updated according to the following formula:
	\begin{equation}\label{rastros-acs}
		\tau_{i,j}(t+1) = (1 - \rho)\cdot \tau_{i,j}(t) + \rho\cdot \tau_{i,j}(0)
	\end{equation} where $\rho \in (0,1]$ is the pheromone decay coefficient. The goal of reducing the intensity of the trail of the edges used by an ant is to diversify the search. With these local updates, the likelihood of two ants walking the same path in the same iteration is reduced. In addition, the \textit{offline} update is carried out with the elitist strategy of the $\mathcal{MM}$AS, that is, only by the ant that has made the best path (the one of minimum cost), according to the formula~(\ref{rastros-mmas}). Either the best path of the last iteration or the best one found so far can be considered.
	
	Another important change from the AS is the calculation of probabilities. In this version, the probability of choosing an edge also depends on a variable $q$ uniformly distributed in [0,1] and a parameter $q_0$. Thus, ants select the next vertex according to the so-called pseudo-random proportional rule: the probability of choosing the vertex $v_{j^*}$ such that
    $j^* = \textit{argmax}_{j\in[1,n]}\left\{\tau_{i,j}\cdot\eta_{i,j}^{\beta}\right\}
    $ (that is, of choosing the vertex at which the product of the number of pheromones and the visibility ---with its relative importance $\beta$--- is maximized) is $q_0$; whereas the probability of choosing according to the standard probabilities of  formula~(\ref{prob-as}) is $1-q_0$.

    These variations in the update of trails and the assignment of probabilities increase the importance of the information collected by the previous ants compared to the exploration of the search space.
\end{subsection}

 \begin{subsection}{Parallelization of the ACO algorithms}\label{aco-par} \label{subsec:parallelACO}

Let us discuss the most common methods in the literature to parallelize these algorithms.
Typically, a system with $n$ processors in parallel is less efficient than a single processor $n$ times faster. However, parallel systems are usually cheaper, so there is a tendency to parallelize processes that were originally sequential. The main reason for parallelizing metaheuristic algorithms such as ACO algorithms is either to increase the quality of the solutions achieved in a given time or, given a required quality, to reach sufficiently good solutions in less time.

When parallelizing ACO algorithms two main strategies can be found in the literature: \textit{fine-grained} parallelization and \textit{coarse-grained} parallelization (\cite{dorigo06}). \textit{Fine-grained} parallelization  consists in assigning very few individuals to each processor so that communication among them is very frequent.  \textit{Coarse-grained} parallelization consists in assigning large subpopulations to each processor with less communication.

It has been empirically proven (\cite{manfrin06}) that the \textit{coarse-grained} is the most efficient strategy for implementing ACO algorithms in parallel. In fact, the version that provides the best results is the independent execution of $n$ copies of the same sequential $\mathcal{MM}$AS
algorithm  with different initializations of the variables.
In this implementation, a maximum number of iterations is set, and copies of the algorithm are executed without any communication between them, choosing the best solution among those provided by the different copies.
This way, the expensive communications performed at the end of all iterations in a typical parallel implementation of ACO are replaced by a single communication phase at the end of the execution of all ACO instances.

A survey on parallelization of swarm intelligence algorithms is presented in~(\cite{mkb2022}). Research works are classified in terms of their parallelization scheme and the kind of hardware they use. Regarding ACO, three types of parallelization are considered in that paper:
\begin{itemize}
    \item {\it Naive} parallelization: it consists in using one thread per ant. It is considered a quite straightforward approach capable of achieving considerable speedups, especially when the dimension of the optimization problem grows.
    \item {\it Fine-grained} parallelization: if there are more cores than ants, then a thread block can be used for each ant, and the threads of this block can be used to perform internal calculations of the ant in parallel. The pheromone deposit can be parallelized, although conflicting updates must be computed sequentially  by using atomic operations.
    Overall, it is considered that the \textit{fine-grained} parallelization improves execution times, although it makes it impossible to have an efficient synchronization method on the hardware level.
    \item {\it Parallel Ant Colonies}: a multi-colony is processed in parallel by assigning each colony to a process. As ACO is an elitist algorithm, only successful ants and their paths  influence the behavior of the colony in the next iteration, so in general most of the work performed by other ants is discarded to follow the successful ant. By using multiple colonies running at the same time, a bigger diversity of successful solutions is produced.
    As each sub-colony is running independently, an eventual synchronization among them is necessary to have the same level of pheromone deposits, although it is recommended to do it only sporadically to avoid extra time.
    \end{itemize}
To cover the different types of parallelization applicable to ACO algorithms, in this paper we will build a process algebra that allows us to specify from a \textit{fine-grained} implementation with almost continuous communication, to the most efficient \textit{coarse-grained} without any exchange of information. In terms of the previous literature, our \textit{fine-grained} scheme will be at some point between the first two parallel schemes from~(\cite{mkb2022}) commented before: each ant will be represented by a process as in the so-called \textit{naive} scheme, although the responsibility to keep the pheromone trail values will be encapsulated by an additional process we will call the {\it graph}. This way, some operations will be performed out of the ant processes themselves ---somehow in the line of the so-called \textit{fine-grained} scheme from that work.\footnote{The purpose of that graph process  will be to enable the updating process of the graph in a simple way, but obviously other representations would be possible (for example, all ants could remotely access a shared variable representing the graph state, the graph state could be distributed among all ants, etc.).}
Besides, our \textit{coarse-grained} scheme will be the most efficient scheme according to~(\cite{manfrin06}) as we mentioned before, that is, executing several completely independent instances of the ACO algorithm and finally getting the best solution found by all of them. However, we will also consider a variant of it where sporadic synchronization among the subpopulations is possible to let them evolve into a common solution, as in the third scheme from~(\cite{mkb2022}) commented before, the so-called \textit{parallel ant colonies} scheme.
\end{subsection}


\section{Process algebra}\label{alg}\label{sec:PA}

Process algebras are mathematical structures that permit expressing the behavior of systems in algebraic terms, what helps to reason formally about their correctness and their properties. Every process algebra consists of a set of operators and a set of atomic actions, which represent indivisible behavior (\cite{fokkink00}, \cite{friso02}). From these operators and actions, a syntax is proposed for the algebraic terms that represent the processes, and their evolution is defined by means of an operational semantics.

There are two types of actions: internal actions and communicating actions. An internal action is an atomic action whose performance cannot be seen by an external observer. It will be denoted by $\mu$.\footnote{In the literature, the usual notation for this internal action is $\tau$. However this notation would produce a conflict with the pheromone trails (also denoted by that symbol) when we formalize the ACO algorithms.} We consider $Act_\mu = Act \cup \{\mu\}$ where $Act$ is the set of communicating actions (we will assume $\alpha \in Act_{\mu}$ so that we can refer to a generic action as $\alpha$ when needed). Communicating actions are related with the sending and reception of information
among processes. In our process algebra, we will consider sending actions and receiving actions, and they will be performed in a synchronous communication.
Besides, we will include probabilistic information in order to represent the behavior of the system when the execution of some actions depends on probability. For a more detailed insight of probabilistic algebras read e.g. (\cite{glabbeek90}, \cite{lopez04}).

We will consider a non-deterministic probabilistic model, that is, both probabilistic and non-deterministic choices are used. In this case, it is usual to consider two different types of transitions: those associated with a probability and those associated with an action. Within this class of model, we can discriminate between alternating models and non-alternating models. In alternating models, there is a strict alternation between probabilistic and non-deterministic choices, while in non-alternating ones we do not have this restriction.

In our case, we will develop a non-deterministic and non-alternating probabilistic model. In these models, it is crucial to determine what kind of states we want to reach after each transition.

\end{subsection}

\begin{subsection}{Process Algebra for ACO, PA$^2$CO}\label{mialg}\label{subsec:PAACO}

In our case, we want to define a probabilistic process algebra that can represent the sending of data between the ants and the graph, as well as the changes made in this data. For this purpose, we will introduce the concepts of state and configuration. As we will see, a state will be a representation of the values assigned to the variables of the process at each moment, which will be modified with the evolution of the process itself. A configuration will be a pair formed by a process and a state (the one associated with its variables).

Before giving the syntax of our process algebra, we will introduce several concepts and some auxiliary definitions.

\begin{definition}[State]
	Let $\Var$ be a set of variables and $\Data$ a set of data. We define a state as any partial function $\sigma: \Var -\rightarrow \Data$.
	The set of all the states will be denoted by $\St$.
\end{definition}
Note that total functions will be denoted by $\longrightarrow$, meanwhile  $-\rightarrow$ will be used to denote partial functions.

These states will be sent by the processes of our process algebra through the communicating actions. Next we will define some functions over the variables of the states:
\begin{definition}[State restricted to a set]
    Let $\sigma \in \St$ be a state and $V \subset \Var$ a subset of variables. We define the state $\sigma$ restricted to $V$ as the state obtained considering only the assignment of $\sigma$ to the variables in $V$, denoted by $\sigma_V$. That is,
    \[	\forall x \in \Var, \ \sigma_{V}(x) = \left\{
	    \begin{array}{ll}
		    \sigma(x) & \textrm{if } x \in V\\
		    \textrm{undefined}    & \textrm{otherwise}\\
	\end{array}
	\right.
	\]
\end{definition}

On the other hand, once processes are evolving, the variables assignment in states will be modified. For that reason, we will consider a set of functions $\Trans$ that update the states.
\begin{definition}[Transformation]
	We define a transformation as a function over states, that is, $\textit{trans}: \St \rightarrow \St$. We denote by $\Trans$ the set of all possible transformations for a set of states.
\end{definition}
The most useful transformations will be those that change a single variable of a state.
\begin{definition}[Single transformation]
Let $\sigma \in \St $, $y \in \Var$, and $d \in \Data$. We define the transformation $[y\mapsto d]:\St\rightarrow \St$ applied to $\sigma$, denoted by $\sigma[y\mapsto d]$, as the transformation where the new state  behaves as $\sigma$ with all the variables but $y$, whose new value will be $d$. That is,
	$$\forall x \in \Var,\
	  \sigma[y \mapsto d](x) = \left\{ \begin{array}{ll}
		\sigma(x) & \textrm{if }  x \neq y\\
		d                 & \textrm{otherwise}\\
	\end{array}
	\right.$$
\end{definition}
When the values of several, but not all, variables are replaced, for the sake of clarity we will consider a composition of several of these single transformations. In the definition of the operational semantics, we will consider a transformation of a state by another state.

\begin{definition}[State transformations]
	Let $\sigma, \sigma' \in \St$ be states. The state transformation of $\sigma'$ applied to $\sigma$ is a  state created by the substitution in $\sigma$ of the assignations given by $\sigma'$, denoted by $\sigma[\sigma']$. That is,
	$$\forall x \in \Var, \
	  \sigma[\sigma'](x) = \left\{ \begin{array}{ll}
		\sigma(x) &   \textrm{if } \sigma'(x) =  \textit{undefined}  \\
		\sigma' (x)  &  \textrm{otherwise}\\
	\end{array}
	\right.$$
\end{definition}

In addition to the transformations, we also consider conditions, which are functions that return Boolean values, i.e. true or false, based on the data assigned to the variables of a given state.

\begin{definition}[Condition]
	Let $\textit{Bool}$ be the set of Boolean values. We define a condition as any function,  $\textit{cond}: \St \rightarrow \textit{Bool}$.
	We denote by $\textit{Cond}$ the set of all the conditions over a set of states.
\end{definition}
Thus, a condition will be true or not at a certain moment in the execution of a process depending on the values of its variables at that time.

\begin{definition}[Actions]\label{def:actions}
    We define the set of actions, denoted by $Act_{\mu}$, as the actions defined by the following BNF expression:
	\[\alpha ::= \mu |\ \textit{trans} |\ c!(\sigma) |\ \recc{\sigma'} \]
    \noindent where $\sigma,\sigma'\in \St$, $\textit{trans} \in \Trans$, and $c$ is a communication channel.
\end{definition}
The action $\mu$ represents the internal action, $\textit{trans}$ represents any transformation over a state, $c!(\sigma)$  is an output communicating action that sends the state $\sigma$ through channel $c$, and $\recc{\sigma'}$ is an input communicating action that receives a state saved in $\sigma'$ through channel $c$. We will write directly $c!$, omitting the parameter, when the full current state is sent.

Once states, transformations, conditions, and actions are defined, we give the syntax to our processes.

		\begin{table}[h]
		
	    \hrulefill
	
	    \centering
	    {\small
	    \begin{tabular}{ll}
            $G ::= N \ |\ P$ &\\
            $N ::= \textit{stop} \ |\ X \ |\ \alpha;G \ |\ [\textit{cond}]G \ |\ \Sigma_{i \in I}N_i \ |\ rec\ X.N$ & (non-deterministic processes)\\
            $P ::= X \ | \ \oplus_{i \in I} [p_i] N_i \ |\ rec\ X.P$ & (probabilistic processes)\\
            \hspace*{-2em}where\\
            $p_i \in [0,1]$ for all $i \in I$ and
            $\Sigma_{i \in I}p_i = 1$ \\
            $\textit{cond} \in \textit{Cond}$ \\
            $X$ is a process variable and
            $I$ is a finite set of indexes \\
            $\alpha\in Act_\mu$
            \end{tabular}
            }
		
		\hrulefill
	    \caption{Syntax of processes.}
	    \label{tab:process}
	    \end{table}

\begin{definition}[Process]
	The syntax of a process is defined as in BNF expression given in Table~\ref{tab:process}, where two types of processes are distinguished,
    \begin{itemize}
    \item the set $\mathcal{P}_N$ of non-deterministic processes and
    \item the set $\mathcal{P}_P$ of probabilistic processes,
    \end{itemize}
    and we consider $N \in \mathcal{P}_N$, $P \in \mathcal{P}_P$, and $G \in \mathcal{P}_N \cup \mathcal{P}_P = \mathcal{P}$.
\end{definition}

 According to the BNF expression for $N \in \mathcal{P}_N$, the non-deterministic processes are:
\begin{itemize}
    \item $\textit{stop}$, the process that does nothing. Although all processes will finish with the action $\textit{stop}$, we will omit it for the sake of clarity. Thus, the process $\alpha;\textit{stop}$ will be represented by $\alpha$,

    \item the process $\alpha;G$ that performs the action $\alpha$ and then behaves like process $G$,
    \item the process $[\textit{cond}]G$ that behaves like $G$ if condition $cond$ holds, and otherwise it cannot evolve (it will be equivalent to the $\textit{stop}$ action in this case),
    \item the recursive process $rec\ X.N$ associated with a non-deterministic process $N$, and
    \item $\Sigma_{i \in I} N_i$, the choice among non-deterministic processes (when applied as a binary operator, it will be denoted by $N_1 + N_2$).
\end{itemize}

In summary, the non-deterministic processes are those that cannot evolve and those that evolve by performing an action from the set $Act_{\mu}$.

On the other hand, we consider as probabilistic processes those in which a probabilistic decision is made ($\oplus_{i \in I}$ $[p_i] N_i$), as well as the  recursive processes associated with a probabilistic process ($rec\ X.P$). As we will see with the operational semantics, these processes evolve by making a probabilistic decision. We will require that, after a probabilistic decision, a non-deterministic process must be executed. Indeed, this will be the desired behavior in the specification of the ACO and avoiding this restriction would, in fact, complicate the operational semantics unnecessarily.

The order of preference for the operators used in the previous definition, from higher to lower precedence, is: the recursive operator, the evaluation of a condition $[\textit{cond}]$, the sequential composition ; , the non-deterministic choice $\Sigma$, and the probabilistic choice $\oplus$.

Note that the recursive process $rec\ X.N$ evolves like the process $N$ when all the occurrences of the variable $X$ in $N$ are replaced by the process $rec\ X.N$. This behavior can be easily understood with a brief example:
\begin{examp} Let $A \in G$ be the process:
\[
A ::= rec X.(\alpha;X)
\]
where $\alpha\in Act_\mu$. If we unfold the recursion once, then the process $A$ can be described as:
\[
A ::=  \alpha; (rec X. (\alpha;X))
\]
which is the process that executes $\alpha$ and repeats $A$ again. Each occurrence of the variable $X$ can be seen as a recursion call of the process $A$, so $A$ is the process that executes $\alpha$ in a loop.
\end{examp}

Let us note that it is allowed to choose among conditions and actions in $\Sigma_{i \in I} N_i$. In fact, we are interested in enabling this behavior in our algebra, as we see in the next example:
\begin{examp}
    Let us suppose that we have a process $Q$ that receives data from $n$ different channels but is interested only in the first data from each input channel. Besides, it wants to send another data when it is requested by the channel $\textit{channelOut}$. Then its behavior can be represented in this way:
\[
Q ::= rec.X(\Sigma_{i = 1}^n [\textit{firstData}_i]\textit{channel}_i?(\sigma);X \+ \textit{channelOut}!;X)
\]

\noindent where the condition $\textit{firstData}_i$ indicates whether it is the first data that is received by the channel $i$ or not.

\end{examp}

The philosophy of this example will be used later when we model the ACO algorithms, in particular when we want the ants that have completed their lap to wait for the others to finish theirs in order to update the traces and continue the search.

Note that the choice operator defines choice in general, although this choice can be a deterministic, external, non-deterministic, or any mixture of them. We will call this choice {\it non-deterministic} because it is so when no additional conditions hold that make it fit into any of the other cases, so it is the most general case indeed.
\begin{examp}
Assuming $val$ is a variable, the choice is deterministic in process $[val=1]a! + [val=0]b!$, since at most one deterministic choice will be available in each case. It is external in process
$a?(x) + b?(x)$ because the choice will be determined by the other processes. It is non-deterministic in process $a! + b!$, as well as in process $[val=1]a!$ $+ [val=1]b!$. Finally, it is some mixture of the previous kinds in process $rec X.([\textit{enough\_air}]\textit{shout\_help}!;X + [\textit{enough\_air}]\textit{shout\_I’m\_drowning}!;X + \textit{waving\_arms}!;X + \textit{receiving\_life\_jacket}?(\sigma') + \textit{receiving\_rubber\_ring}?(\sigma'))$).
\end{examp}

\end{subsection}

\begin{subsection}{Parallel composition in two steps}

Once the operators, the set $\St$ of states and the syntax of our processes are described, we define the configurations. Before the formal definition, a main intuition of the concept is given. A configuration is a pair $\ll G,\sigma \gg$ with $G \in \mathcal{P}$ and $\sigma \in \St$ so that, intuitively, $\ll G,\sigma \gg$ indicates that the process $G$ is going to be executed from the state of its variables $\sigma$. As we will see, in the specifications of the ACO algorithms several processes will be executed in parallel, and we will be interested in making each of them modify its variables locally. Thus, each process has its own state, independent of the others.
Besides we will avoid, by the construction of the operational semantics, any competition between the choices offered by a probabilistic process and the choices offered by a non-deterministic process being in parallel with it (note that the former quantifies the likelihood of its choices but the latter does not, so they cannot directly compete). Due to the non-deterministic nature of the parallel composition, the 
composition of processes of both kinds together will be neither pure probabilistic nor pure non-deterministic.

To avoid this ambiguity, the operational semantics will divide the execution in parallel into two steps:
\begin{itemize}
    \item First, it is non-deterministically decided which one of the different configurations "wins the race", that is, which one is going to be executed next.
    \item Second, if the chosen one is a probabilistic configuration, then a probabilistic decision is made next; and if the chosen one is non-deterministic, then an action will be executed next.
\end{itemize} After performing these two steps, we will have a new parallel configuration and we will repeat the choice and execution process again. Accordingly, we will consider that all configurations of the form $\|_{i\in I}C_i$ are {\it non-deterministic}, and they will lead us with an internal action to a configuration of the form $\|_{i\in I}^j C_i$, indicating that $C_j$ with $j\in I$ has won the race. In this way, the (probabilistic or non-deterministic) nature of the configuration $\|_{i\in I}^j C_i$ will depend on the nature of the process present in $C_j$.

\begin{definition}[Configuration]
	We define the syntax of the configurations using the following BNF expression:
	$$ C ::= \ll G,\sigma \gg\ |\ \|_{i\in I} C_i \ |\ \|_{i\in I}^j C_i$$
	where $G \in \mathcal{P}$, $\sigma \in \St$, $I$ is a finite set of indices, and $j \in I$.
\end{definition}

In Table~\ref{SEOP} the operational semantics of the configurations is presented. We denote by $\alpha^*$ the actions unrelated to sending data, i.e. $\alpha^* \in \Trans \cup \{\mu\}$.

\begin{table}
   \begin{tabular}{ll}
		\hline
		\\[-0.5em]
		\footnotesize{$R1$} $\frac{}{\ll \alpha;G,\sigma\gg \tran{\alpha} \ll G,\sigma\gg }$
	     {\footnotesize{$\alpha \in\{c!,\mu\} $}}  &
		\footnotesize{$R2$} ${\frac{}{\ll \recc{\sigma'};G,\sigma\gg \tran{\recc{\sigma'}} \ll G,\sigma[\sigma']\gg }}$\\[2em]
		\multicolumn{2}{c}{\footnotesize{$R3$} ${ \frac{}{\ll \textit{trans};G, \sigma\gg \tran{\textit{trans}} \ll G,\textit{trans}(\sigma)\gg }}$ }\\[2em]
        \footnotesize{$R4$} ${ \frac{\textit{cond}(\sigma),\; \ll N,\sigma \gg \tran{\alpha^*} \ll N',\sigma'\gg}{\ll [\textit{cond}]N,\sigma\gg \tran{\alpha^*} \ll N',\sigma'\gg }}$ \footnotesize{$\alpha^* \!\in\! \Trans\! \cup\! Act_{\mu}$}&	
		\footnotesize{$R4'$} ${ \frac{\textit{cond}(\sigma)}{\ll [\textit{cond}]P,\sigma\gg \tran{\mu} \ll P,\sigma\gg }}$  \\[2em]
		\footnotesize{$R5$} ${ \frac{ \ll N_j, \sigma\gg \tran{\alpha_j} \ll N'_j,\sigma_j\gg}
		{\ll \Sigma_{i \in I} N_i, \sigma\gg \tran{\alpha_j} \ll N'_j,\sigma_j\gg }}$ {\footnotesize $j\! \in\! I$}  &	
		\footnotesize{$R6$} $\frac{}{\ll \oplus_{i \in I}[p_i]N_i, \sigma\gg \tranp{\ }{p_j} \ll N_j, \sigma\gg}$\ \footnotesize{$j\! \in\! I$}\\[2em]
		\footnotesize{$R7$} $\frac{\ll G_j,\sigma_j \gg \tran{\alpha^*} \ll G'_j,\sigma'_j \gg}
		{\|_{i\in I} \ll G_i,\sigma_i\gg \tran{\mu} \|_{i \in I}^j \ll G_i,\sigma_i \gg}$
		\footnotesize{$\alpha^* \!\in\! \Trans\! \cup\! \{\mu\}$} &
		\footnotesize{$R8$} ${ \frac{ \ll G_j,\sigma_j \gg \tranp{\ }{p} \ll G'_j,\sigma'_j\gg }{\|_{i\in I} \ll G_i,\sigma_i\gg \tran{\mu} \|_{i \in I}^j \ll G_i,\sigma_i\gg}}$ \\[2em]
		\multicolumn{2}{c}{\footnotesize{$R9$} ${ \frac{ \ll G_j,\sigma_j\gg \tran{\recc{\sigma}} \ll G'_j,\sigma'_j\gg \hspace{4mm} \ll G_k,\sigma_k\gg \tran{c!} \ll G'_k,\sigma'_k \gg}{\|_{i\in I} \ll G_i,\sigma_i\gg \tran{\mu} \|_{i \in I}^j \ll G_i,\sigma_i\gg}}$ $j,k \in I$} \\[2em]
		\multicolumn{2}{c}{\footnotesize{$R10$} ${ \frac{\ll G_j,\sigma_j\gg  \tran{c!} \ll G'_j,\sigma'_j\gg  \hspace{4mm} \ll G_k,\sigma_k\gg  \tran{\recc{\sigma}} \ll G'_k,\sigma'_k\gg }{\|_{i\in I} \ll G_i,\sigma_i\gg  \tran{\mu} \|_{i \in I}^j \ll G_i,\sigma_i\gg }}$  $j,k \in I$} \\[2em]
		\multicolumn{2}{c}{\footnotesize{$R11$}
		${ \frac{\ll G_j,\sigma_j\gg  \longrightarrow_p \ll G'_j,\sigma'_j\gg }{\|_{i\in I}^j \ll G_i,\sigma_i\gg  \longrightarrow_p \|_{i \in I} \ll G'_i,\sigma'_i\gg }}$\ $\space$ {\footnotesize $\ll G'_i,\sigma'_i\gg  = \ll G_i, \sigma_i\gg  \forall i \neq j$}\ } \\[2em]
		\multicolumn{2}{c}{\footnotesize{$R12$}
		$\frac{\ll G_j,\sigma_j\gg  \tran{\alpha^*} \ll G'_j,\sigma'_j\gg }
		      {\|_{i\in I}^j \ll G_i,\sigma_i\gg  \tran{\alpha^*} \|_{i \in I} \ll G'_i,\sigma'_i\gg }$\ 
		      \footnotesize{$\left .
		        \begin{array}{c}
		            \ll G'_i,\sigma'_i\gg  = \ll G_i, \sigma_i\gg \\
		            \forall i \neq j, \alpha^* \in \Trans \cup \{\mu\}
		        \end{array}
		        \right . $}}
		\\[2em]
		\multicolumn{2}{c}
		{\footnotesize{$R13$}
		  ${ \frac{\ll G_j,\sigma_j\gg  \tran{c!} \ll G'_j,\sigma'_j\gg  \hspace{4mm} \ll G_k,\sigma_k\gg  \tran{\recc{\sigma}} \ll                G'_k,\sigma'_k\gg }
		  {\|_{i\in I}^j \ll G_i,\sigma_i\gg  \tran{\mu} \|_{i \in I} \ll G'_i,\sigma'_i\gg }}$\ 
		  \footnotesize{$\left . \begin{array}{c}
		                    \ll G'_i,\sigma'_i\gg  = \ll G_i, \sigma_i\gg \\
		                    \forall i \notin \{j,k\} \subseteq I
		                    \end{array} \right . $}} \\[2em]
		\multicolumn{2}{c}{\footnotesize{$R14$}
		  ${ \frac{\ll G_j,\sigma_j\gg  \tran{\recc{\sigma}} \ll G'_j,\sigma'_j\gg  \hspace{4mm} \ll G_k,\sigma_k\gg  \tran{c!} \ll G'_k,\sigma'_k\gg }{\|_{i\in I}^j \ll G_i,\sigma_i\gg  \tran{\mu} \|_{i \in I} \ll G'_i,\sigma'_i\gg }}$\ 
		  {\footnotesize $\left . \begin{array}{c}
		                     \ll G'_i,\sigma'_i\gg  = \ll G_i, \sigma_i\gg  \\
		                     \forall i \notin \{j,k\} \subseteq I
		                     \end{array} \right .$}\ }\\[2em]
		\footnotesize{$R15$} ${ \frac{\ll N,\sigma\gg  \tran{\alpha} \ll G,\sigma'\gg }{\ll rec\ X.N,\sigma\gg  \tran{\alpha} \ll G[rec\ X.N/X], \sigma'\gg }}$\  &
		\footnotesize{$R16$} ${ \frac{\ll P,\sigma\gg  \longrightarrow_p \ll G,\sigma'\gg }{\ll rec\ X.P,\sigma\gg  \longrightarrow_p \ll G[rec\ X.P/X], \sigma'\gg }}$ \\[2em]
		\hline
	\end{tabular}
 \caption{Operational Semantics of the Process Algebra.}\label{SEOP}%
\end{table}

\begin{itemize}
    \item
Rule R1 indicates that the sending of states and the internal action are executed without changing the state.
\item
Rule R2 states that if the process receives a state, then we consider the state transformation with the received state applied to the initial state.
\item Rule R3 indicates that the transformations are applied in the current state, and the new state is precisely the result of the transformation.
\item Rule R4 states that if a condition applied to the state holds, and it is followed by a non-deterministic process that can evolve performing a transformation or a communicating action, then the whole process evolves by performing that action. Rule R4' indicates that if a condition is fulfilled and it is followed by a probabilistic process, then the whole process evolves by performing an internal action {\it without} performing the probabilistic action yet (in this way, we avoid simultaneously enabling non-deterministic and probabilistic steps from the same configuration). In both rules, if the condition is not fulfilled then the process will not evolve.
\item Rule R5 is the usual one for non-deterministic choice operators: an action is non-deterministically chosen to be performed. A process having multiple choices can evolve only like any of the processes denoting these choices can evolve themselves. Hence, if one of these processes starts with a condition which does not hold then the process making the choice cannot get stuck by choosing to become it, as that process cannot evolve.\footnote{Note that if rule R4 allowed to just ``remove'' the $[cond]$ prefix in a process $[cond]N$ when $cond$ holds (instead of making $N$ {\it evolve} into $N'$ as it actually does), then, by applying rule R5, a process $R::=[cond]a?(x) + [cond]b?(x)$ could evolve in a step into $Q::=a(x)?$ or $Q'::=b(x)?$ when $cond$ holds. In particular, if that process $R$ were composed in parallel with a process $a!$, then this system would be blocked if process $R$ evolved {\it in particular} into $Q'$: process $a!$ can only send a message through channel $a$, whereas $Q'$ can only receive a message through channel $b$. The actual rule R4 avoids this problem, as process $R$ cannot {\it evolve} into $Q$ or $Q'$. Since R5 allows any evolution that is allowed to any of its choices, by the application of R4 to these choices process $R$ can {\it perform} as $Q$ would do, or perform as $Q'$ would do (which is quite different to {\it becoming} $Q$ or $Q'$ itself). This way, $R$ {\it simultaneously} enables both choices $a?(x)$ and $b?(x)$. In, particular, note that process $R$ is equivalent to $[cond](a?(x) + b?(x))$.}
\item Rule R6 is the usual one for probabilistic choice operators.
The process evolves deciding probabilistically.
\item Rules R7-R10 represent the first step of the parallel composition. In this step, a process of the parallel composition `wins the race' to be the one performed in the second step. To do so, only a configuration whose associated process can move forward can win the race. For that reason, in R7 it is needed that the $j$ configuration evolves by a transformation or an internal action. In R8, the winning configuration is able to evolve with a probabilistic transition. Finally, in R9 and R10 the $j$ configuration must be able to participate in a communication, either as the sender or as the receiver. Since communications are synchronous, both the sender party and the receiver party must be ready to communicate. Hence, configuration $j$ can win the race only if there exists a configuration~$k$ it will be able to perform a communication with, where that configuration $k$ plays the opposite communicating role.
\item Rules R11-14  represent the second step of the parallel composition. In all of them we start from a parallel configuration in which the configuration that will be executed first has already been chosen. If a probabilistic configuration has won the race, then a probabilistic decision is executed and we return to a non-deterministic parallel configuration waiting for the next configuration to be executed to be selected (R11). If a non-deterministic configuration has won the race, then we must distinguish whether it is an action of sending or receiving states or not. If it is neither of these two actions, then the configuration executes the action and, as before, returns to a parallel configuration waiting for the next configuration to run to be selected (R12). On the other hand, if it is an action of sending or receiving states then, in order to be executed, the inverse action must also be executed. In this case, both the configuration that 'has won the race' and the one associated with the process that executes the inverse communication action will evolve (R13 and R14).
\item In rule R15, the evolution of a recursive non-deterministic process is defined. If the process $N$ evolves into $G$ by performing some action $\alpha$, then the recursive process $(rec X.N)$ will evolve into $G[rec X.N/X]$ executing that action. That is, it evolves, performing the action $\alpha$, into the process $G$ where all the occurrences of the variable $X$ are replaced by the process $(rec X.N)$.
\item In rule R16, the evolution of a recursive probabilistic process is defined. If the process $P$ evolves into $G$ by performing a probabilistic action, then the recursive process $(rec X.P)$ will evolve into $G[rec X.P/X]$ with the same meaning as before. %
\end{itemize}
\end{subsection}

\section{Formalization of the \textit{fine-grained} parallelization of ACO}\label{fine}\label{sec:fine-grained}
	
	Let us formalize the ACO algorithms implemented in parallel. Both the \textit{fine-grained} and \textit{coarse-grained} implementations will be specified with PA$^2$CO. In this section we will present the \textit{fine-grained} approach.
	
	We will start by specifying two versions of the AS: a first version with a number $m$ of processes representing each one an ant and another process for the graph, all of them executed in parallel so that each ant completes one path in each iteration; and a second version, modifying the previous one so that the same ant can perform several paths in the same iteration, in an attempt to improve the efficiency of the implementation when some ants are faster.
    This second approach would not be very useful in a fully homogeneous environment like, for instance,  a typical GPU-based implementation. However, it would be suitable  in a distributed or cloud computing implementation of ACO with much more heterogeneous environments, where processors with very different speeds and memory availability would be expected. In fact, this might be the only available approach when the desired parallelization level exceeds the number of processors of any parallel computer one can afford to use.

    Afterwards, we will discuss the specification of the $\mathcal{MM}$AS which, as we will see, differs very little from those of the AS. Although the difference between these versions is small, it is interesting to refer to both of them since the AS was the first ACO algorithm to be considered and explains neatly how these algorithms work, whereas the \mmas~ is the most used version of ACO. To conclude this section, we will introduce the ACS specification in the same two cases as in the case of AS.
	
    In the rest of the section, we will consider that the graph of the TSP has $n$ vertices, $I = \{1,\cdots,n\}$, $\Data = \mathbb{R^+} \cup \textit{List}(\mathbb{R^+}) \cup \textit{List(List}(\real^+))$ (where $\textit{List}(A)$ denotes the set of the lists of elements of $A$), and $0\in \mathbb{R^+}$.

\begin{subsection}{Formalization of fine-grained AS with $m$ ants}\label{form-as-m}
	We start modelling the AS algorithm. Let $m \geq 1$ be the number of ants. We will consider two types of processes  for this \textit{fine-grained} parallelization of AS: \begin{itemize}
	    \item one describing the behavior of an ant (there will be $m$ processes of this kind), and
     \item the other one describing the behavior of the graph.
	\end{itemize}
	In Table~\ref{tab:setsVTC1} the sets of variables, transformations, and conditions needed are shown. They are the base from where the rest of the cases will be built, and in the formalization of each ACO algorithm, we will reuse most of them. The new ones of the following versions will appear in bold letters in the corresponding table of variables, transformations, and conditions.
		
		\begin{table}[h]
		
	    \hrulefill
	
	    \centering
	    {
	    \begin{tabular}{l}
		$\Var = V_1 \cup V_2 \cup V_3$ where:\\
		\begin{tabular}{l}
			$V_1 = \{\textit{numIt, waiting, final, notified}\}$ \\
			$V_2 = \{\textit{path}_k, \textit{edges}_k, L_k | 1 \leq k \leq m \}$ \\
			$V_3 = \{\tau_{i,j} | i,j \in I, i \neq j\}$
		\end{tabular} \\
		$\Trans = T_1 \cup T_2 \cup T_3$ where:\\
		\begin{tabular}{l}
			$ T_1 = \{\textit{newIt, trailUpd, reset, unblock, finish}\}$\\
			$T_2 = \{\textit{block}_k, \textit{notify}_k | 1 \leq k \leq m\}$\\
			$T_3 = \{\textit{move}_{i,j}^k | 1 \leq k \leq m, i,j \in I, i \neq j\}$\\
		 \end{tabular} \\
		$Cond =  C_1 \cup C_2  $ where:\\
		\begin{tabular}{l}
			$C_1 = \{\textit{end, end}_A\textit{, completeIt, allNotif}\}$ \\
			$C_2 = \{\textit{complete}_k, \textit{wait}_k, \textit{notified}_k | 1 \leq k \leq m\}$ \\
		\end{tabular}
		\end{tabular}}
		
		\hrulefill
	    \caption{Sets of variables, transformations and conditions of AS with m ants.}
	    \label{tab:setsVTC1}
	    \end{table}
		
The variable $\textit{numIt}\in \real^+$ will keep the number of iterations completed in the algorithm so far. We will consider that an iteration ends when all the ants have completed a Hamiltonian path. In our modelling, the graph will block the ants that have completed a lap, so that they wait for the others. Thus, the traces are updated only after they all have finished. To be able to specify these blockings, we enter the variable $\textit{waiting}\in \textit{List}(\real^+)$, which will carry the list of ants blocked at each moment, and the transformation $\textit{block}_k$, that will include the ant $a_k$ into the waiting list. The variable $\textit{final}\in \{0,1\}$ will indicate if the algorithm has ended with~1, and if it has not with~0. The graph will send this information to the ants so that they stop their search. The variable $\textit{notified}\in \textit{List}(\real^+)$ will store the list of ants that have received this information.
		In addition, for each $k \in \{1,\cdots,m\}$ we introduce the variables $\textit{path}_k \in \textit{List}(\real^+), \textit{edges}_k \in \textit{List}(\real^+)$ and $L_k\in \real^+$. The variable $\textit{path}_k$ will keep the list of vertices that the $k$-th ant has already visited in the current iteration, $\textit{edges}_k$ will save the list of edges already chosen by that ant, and $L_k\in \real^+$ will indicate the length (or cost) of its lap so far. Although the information in $\textit{edges}_k$ can be obtained from that of $\textit{path}_k$, we separate them for the sake of clarity.
		Finally, $ \forall i,j \in I, i\neq j$ we consider the variable $\tau_{i,j}\in \real^+$ that will save the amount of pheromone in the edge $e_{i,j}$ of the graph.

	    \begin{table}[h]
	    \centering
	    \hrulefill
	    {
	    \[\begin{array}{r l}
	        \textit{newIt}: \St &\longrightarrow \St\\
			\sigma &\longrightarrow \sigma[\textit{numIt} \mapsto \textit{numIt}\+ 1]\\
			\textit{trailUpd}: \St &\longrightarrow \St\\
			\sigma &\longrightarrow \sigma[\tau_{i,j} \mapsto r(\tau_{i,j})] \\[0.5em]
			\multicolumn{2}{c}{r(\tau_{i,j}) = (1-\rho)\cdot \tau_{i,j} + \Sigma_{k = 1}^m\Delta \tau_{i,j}^k \textrm{\ where \ }
		    \Delta \tau_{i,j}^k = \left\{ \begin{array}{ll}
			    \frac{Q}{L_k}&   \textrm{if } e_{i,j} \in \textit{edges}_k \\
			    0  &   \textrm{otherwise}\\
		    \end{array}
		    \right.} \\[0.5em]
			\textit{reset}: \St &\longrightarrow \St\\
			\sigma &\longrightarrow \sigma[\overline{\sigma}]  \\
                &\begin{array}{clll}
                    \textrm{ where }\hspace*{2em} \overline{\sigma}: & \Var &-\rightarrow \Data \\
                    & \textit{path}_k & \longrightarrow\textit{last(path}_k) & \hspace*{2em} \forall k \in \{1,\cdots,m\} \\
                    & \textit{edges}_k &\longrightarrow [~] &\hspace*{2em}\forall k \in \{1,\cdots,m\} \\
                    & L_k & \longrightarrow 0 &\hspace*{2em} \forall k \in \{1,\cdots,m\} \\[0.5em]
                \end{array}\\[0.5em]
			\textit{unblock}: \St &\longrightarrow \St \\
			\sigma &\longrightarrow \sigma[\textit{waiting} \mapsto [~]] \\
			\textit{block}_k: \St &\longrightarrow \St\\
    		\sigma &\longrightarrow \sigma[\textit{waiting} \mapsto \textit{waiting}\!:\!k]
      \\
			\textit{move}_{i,j}^k: \St &\longrightarrow \St\\
			\sigma &\longrightarrow \sigma[\textit{path}_k \mapsto \textit{path}_k\!:\!v_j][\textit{edges}_k \mapsto \textit{edges}\!:\!e_{i,j}][L_k \mapsto L_k + d_{i,j}] \\
			\textit{finish}: \St &\longrightarrow \St\\
			\sigma &\longrightarrow \sigma[\textit{final} \mapsto 1] \\
			\textit{notify}_k: \St &\longrightarrow \St\\
			\sigma &\longrightarrow \sigma[\textit{notified} \mapsto \textit{notified}\!:\!k] \\
	    \end{array} \]}
	    \hrulefill
	    \caption{Transformations of AS with $m$ ants.}
	    \label{tab:trans1}
	    \end{table}
	
	    The definition of the transformations is given in Table~\ref{tab:trans1}.
	    After each iteration of the algorithm, the transformation $\textit{newIt}$ will increase the counter $\textit{numIt}$ and the transformation $\textit{trailUpd}$ will update the amount of pheromone in the edges of the graph. Thus, the function $r$ updates the traces according to the formula~(\ref{rastros-as}) of  Section~\ref{aco-as}, $\rho$ is the decay coefficient of the pheromones, and $Q$ is a constant provided by the programmer.
		When the next iteration is about to start, we will use the transformations $\textit{unblock}$ to unblock all the ants and $\textit{reset}$ to restart the variables $\textit{path}_k$, $\textit{edges}_k$, and $L_k$ $\forall k\in\{1,\cdots,m\}$. Note that, in the remainder of this text, we will write $\textit{last(l)}$ to denote the last element of the list $\textit{l}$ and $\textit{l:e}$ to denote the list $l$ adding the element $e$ in the end.\label{cambio1}
		
		Once the iteration has started, we will use $\textit{move}_{i,j}^k$ to move the k-th ant, $a_k$, through the edge $e_{i,j}$ with $i\neq j$. When the k-th ant has finished a lap, the graph will block it and put it into the $\textit{waiting}$  list with the transformation $\textit{block}_k$.
	    Finally, when the algorithm has finished, the graph will update the value of $\textit{final}$ to $1$ with the transformation $\textit{finish}$ sending this information to each ant, and saving it in the list $\textit{notified}$ with the transformation $\textit{notify}_k$, which adds the k-th ant to the list of notified ants.

        \begin{table}[h]
	    \centering
	    \hrulefill
	    {\tiny

	    \[
     \begin{array}{l c l}
		    \textit{complete}_k(\sigma) =\left\{ \begin{array}{ll}
			    \textit{True} &   \textrm{if } \textit{len}(\sigma(\textit{path}_k)) = n \\
			    \textit{False} &    \textrm{otherwise}\\
                \end{array} \right .
                & \hspace*{6em} &

		    \textit{wait}_k(\sigma) = \left\{ \begin{array}{ll}
			    \textit{True}  &  \textrm{if } k \in \sigma(\textit{waiting}) \\
			    \textit{False} &   \textrm{otherwise}\\
                \end{array} \right . \\[1em]

                \textit{completeIt}(\sigma) = \left\{ \begin{array}{ll}
			    \textit{True}  &  \textrm{if } \textit{len}(\sigma(\textit{waiting})) = m \\
			    \textit{False} &   \textrm{otherwise}\\
                \end{array} \right .&\hspace{2em} &

                \textit{end}_A(\sigma) = \left\{ \begin{array}{ll}
			    \textit{True}  &  \textrm{if } \sigma(\textit{final}) = 1 \\
			    \textit{False} &   \textrm{otherwise}\\
                \end{array} \right . \\[1em]

                \textit{notified}_k(\sigma) = \left\{ \begin{array}{ll}
			    \textit{True}  &  \textrm{if } k \in \sigma(\textit{notified})  \\
			    \textit{False} &   \textrm{otherwise}\\
                \end{array} \right . &\hspace{2em} &

                \textit{allNotif}(\sigma) = \left\{ \begin{array}{ll}
			    \textit{True}  &  \textrm{if } \textit{len}(\sigma(\textit{notified})) = m \\
			    \textit{False} &   \textrm{otherwise}\\
                \end{array} \right .\\\\[1em]

                \multicolumn{3}{l}{\textit{end}(\sigma) = \left\{ \begin{array}{ll}
			    \textit{True}  &  \textrm{if }  (\sigma(\textit{numIt}) = \textit{maxIt}) \textrm{ or } (\textit{len}(\sigma(\textit{waiting})) = m        \textrm{ and } \sigma(\textit{path}_k) = \sigma(path_{k+1}) ~ \forall k \in \{1,\cdots,m-1\})\\
                    \textit{False} &   \textrm{otherwise}\\
                \end{array} \right .} \\
            \end{array}
                \]}

        $\forall \sigma \in \St$ and $\forall k \in \{1,\cdots,m\}$

        \hrulefill
	    \caption{Conditions of AS with $m$ ants.}
	    \label{tab:cond1}
	    \end{table}

             The definition of the conditions is given in Table~\ref{tab:cond1}.
     The condition $\textit{complete}_k$ will tell us whether the k-th ant has completed a lap or not, that is, whether the length of $\textit{path}_k$ matches the number of vertices $n$. The condition $\textit{wait}_k$ will say if the k-th ant is blocked by the graph or not, that is, if the k-th ant is in the $\textit{waiting}$ list, for each $k \in \{1,\cdots,m\}$. The condition $\textit{completeIt}$ will be true when all the ants have completed one lap, that is, when the length of the $\textit{waiting}$ list is $m$.
		
		The condition $\textit{end}$ will be true if one of these two options is fulfilled:
		\begin{enumerate}
			\item $\textit{numIt}$ matches the maximum of iterations, or
			\item the length of $\textit{waiting}$ is $m$ and $\textit{path}_k = \textit{path}_{k+1} ~\forall k \in \{1,\cdots,m-1\}$.
		\end{enumerate}
		 This condition will tell the graph that the algorithm has finished so that it can change the value of the variable $\textit{final}$ to 1 and send it to the ants. Then, the ants will evaluate the condition $\textit{end}_A$, that will be true when the variable $\textit{final}$ is 1. The condition $\textit{notified}_k$ will be true when the k-th ant has received the new value of $\textit{final}$, that is, it is in the list $\textit{notified}$. Finally, $\textit{allNotif}$ will be true when all the ants have been notified.

		The channel of communication between an ant $k$ and the graph will be denoted by $\textit{grA}_k$, that is, we will consider $m$ communication channels.
        Note that PA$^2$CO allows to have several simultaneous sending and receiving actions in the same channel. However, we use one channel per ant so that it is easier to formalize which $\textit{block}_k$ transformation needs to be performed by the graph after receiving the information of an ant. Also, it should be noted that, in this specification and in the rest of the versions, the processes will send complete states, so we will simply write $c!$ to indicate that the current state is sent through the channel $c$ (omitting the parameter).
		
		Once we have defined the sets of variables, transformations, and conditions, the processes for the ants and the graph can be defined as given in Table~\ref{tab:PA1}.
		\begin{table}[h]
	    \centering
	    \hrulefill
	    {
	    \vspace*{0.5em}
	
			k-th ANT
		\[\begin{array}{l}
		     A^k = rec\ X.(\textit{$grA_k$}?(x);([\textit{end}_A]\textit{stop} + [\neg \textit{end}_A]A_{\textit{info}}^k))\\
		     A_{info}^k = recY.(\oplus_{i,j \in I} [p_{i,j}^k]\textit{move}_{i,j}^k;([\textit{complete}_k]A_C^k \+ [\neg \textit{complete}_k]Y))\\
		     A_C^k = grA_k!;X
		\end{array} \]
	    	GRAPH
		\[\begin{array}{l}
		    G = rec\ X.(\Sigma_{k = 1}^{m}[\neg \textit{wait}_k]\textit{grA}_k!;X  +  \Sigma_{k = 1}^m \textit{grA}_k?(x);\textit{block}_k;X + [\textit{completeIt}]G_C)\\
		    G_C = \textit{trailUpd};\textit{newIt}; ([\textit{end}]\textit{finish};G_F \+ [\neg \textit{end}]\textit{unblock};\textit{reset};X)\\
		    G_F = recY.(\Sigma_{k = 1}^m [\neg \textit{notified}_k]grA_k!;\textit{notify}_k;Y \+ [\textit{allNotif}]\textit{stop})\\
		    \end{array}\]
		}
	    \hrulefill
	    \caption{Process algebra of AS with m ants.}
	    \label{tab:PA1}
	    \end{table}

    Thus, the process representing each ant $k\in\{1,\cdots,m\}$ is a recursive process that begins by asking the graph for information to get the amount of pheromone of each edge, $\tau_{i,j}$, and evaluates the condition $\textit{end}_A$. If the algorithm is finished, the process ends. Otherwise, the process $A^k_{\textit{info}}$ is performed and it is decided which vertex to move to by calculating the probabilities $p_{i,j}^k$ according to the information obtained from the graph and the formula of the AS. Once the next edge, $e_{i,j}$, has been selected, the process updates the lists $\textit{path}_k$ and $\textit{edges}_k$ and the variable $L_k$ with the transformation $\textit{move}_{i,j}^k$. At that point, the process evaluates the condition $\textit{complete}_k$. If it has finished the lap, then it sends the information to the graph and restarts the recursive process that represents the ant. If, on the other hand, the lap has not been completed yet, then it goes back to the recursive process $A_{\textit{info}}^k$ to select the next edge.

    The process that represents the graph must manage the information of all the ants with the blockings as well as the end of the process. It is, therefore, a recursive process able to perform three different options:
    \begin{enumerate}
	\item to send its information to any ant that requests it that it is not blocked, that is, that has not        completed a lap in this iteration. Then it restarts the recursive process.
        \item to receive information from one of the ants, in which case it blocks this ant and restarts the recursive process.
        \item to evaluate the condition $\textit{completeIt}$. If true, then it means that all the ants have completed their lap, so the traces are updated and the transformation $\textit{newIt}$ is executed  to update the counter $\textit{numIt}$. The condition $\textit{end}$ is then evaluated  to find out whether the algorithm is over or not. If it is over, then the process runs the transformation $\textit{finish}$ to change the value of the variable $\textit{final}$ to 1 and execute the process $G_F$. If not, then all the ants are unlocked, restarted and returned to the beginning of the recursive process.	
    \end{enumerate}
		
    Note that, if $\textit{completeIt}$ is true, then neither option 1 nor option 2 will be true. Option 1 cannot be true because all ants will be blocked, so the condition $\neg \textit{wait}_k$ will be false for all $k \in \{1,\cdots,m\}$. Option 2 will also not be possible because, if all ants are blocked, then none can send information. Thus, when an iteration has been completed, only option 3 can be executed.
		
    Finally, the process $G_F$ is a recursive process whose goal is to notify all the ants that the algorithm has finished. To do this, it sends the graph data to the ants (with the new value of the variable $\textit{final}$) and when all the ants have been notified, the process ends.
		
    With all this, assuming $M = \{1,\cdots,m\}$, we can model the AS as: $$\|_{k \in M} \ll A_k,\sigma_k \gg \| \ll G,\sigma_G \gg $$ where $\forall k \in M$ we have:
		\[{
		    \begin{array}{clcl c clcl}
		        \sigma_k : &\Var &-\rightarrow &\Data & & \sigma_G: &\Var &-\rightarrow&\Data\\
			    &\textit{path}_k &\longrightarrow  &[v_{0,k}] & & &\textit{numIt} &\longrightarrow  &0\\
			    &\textit{edges}_k &\longrightarrow  &[~] & & &\textit{final} &\longrightarrow  &0\\
			    &L_k &\longrightarrow  &0 & & &\textit{notified} &\longrightarrow  &[~]\\
                & & & & & &\textit{waiting} &\longrightarrow  &[~]\\
			    &   &             &  & & &\tau_{i,j} &\longrightarrow  &\tau_0 ~\forall i,j\! \in\! \{1,\cdots,n\}, i\neq j
		    \end{array} } \]
    where $v_{0,k}$ is the vertex at which the k-th ant is placed at the beginning of the algorithm and $\tau_0$ is the initial amount of pheromone on each edge. That is, $\sigma_k$ and $\sigma_G$ are the states with the initialized values of the variables.
\end{subsection}

\begin{subsection}{Formalization of fine-grained AS with $m$ free ants}\label{form-as-ml}
	
	In this second version of the AS we do not force ants to wait for each other. In this way, if an ant is assigned to a faster processor, then we take advantage of this speed instead of making the ant wait for the other ants assigned to slower processors to finish their laps. In order to preserve the philosophy of the AS, we will continue to consider $m$ paths to carry out the updating of traces, although several of them correspond to the same ant.
	
	As the ants will not be blocked any more, we eliminate all the variables, transformations and conditions related to the blockages introduced in the previous section, leaving the sets of variables, transformations and conditions of Table~\ref{tab:setsVTC2} with the new transformations and conditions defined in Table~\ref{tab:trans2} and Table~\ref{tab:cond2}, respectively. Note that we write \textit{list1 ++ list2} to denote the list resulting from joining lists \textit{list1} and \textit{list2} in that order.\label{cambio2}

		\begin{table}[h]
	    \centering
	    \hrulefill
	
	    {
	    \begin{tabular}{l}
        $\Var = V_1 \cup V_2 \cup V_3$ where:\\
	    \begin{tabular}{l}
		    $V_1 = \{\textit{numIt, final, \textbf{paths}, \textbf{edges}, \textbf{costs}, notified, \textbf{infoSent}}\}$\\
		    $V_2 = \{\textit{path}_k, \textit{edges}_k, L_k  \ |\   1 \leq k \leq m \}$\\
		    $V_3 = \{\tau_{i,j} \ |\   i,j \in I, i \neq j\}$\\
	    \end{tabular} \\
	    $\Trans =  T_ 1 \cup T_2 \cup T_3$ where: \\
	    \begin{tabular}{l}
		    $T_1 = \{\textit{\textit{newIt, trailUpd, finish, \textbf{resetLoop}, \textbf{infoAdded}}}\}$\\
		    $T_2 = \{\textit{\textbf{reset}}_k, \textit{notify}_k, \textit{\textbf{addPath}}_k  \ |\   1 \leq k \leq m\}$ \\
		    $T_3 = \{\textit{move}_{i,j}^k  \ |\   1 \leq k \leq m, i,j \in I, i \neq j\}$ \\
	    \end{tabular} \\
	    $Cond = C_1 \cup C_2 $ where: \\
	    \begin{tabular}{l}
		    $C_1 = \{\textit{end, end}_A\textit{, completeIt, allNotif, \textbf{maxInfo}}\}$ \\
		    $C_2 = \{\textit{complete}_k, \textit{notified}_k \ |\  1 \leq k \leq m\}$\\
	    \end{tabular}
		\end{tabular}}
		
		\hrulefill
	    \caption{Sets of variables, transformations, and conditions for AS with m free ants.}
	    \label{tab:setsVTC2}
	    \end{table}	
	
		\begin{table}[h]
	    \centering
	    \hrulefill
	
	    {
	    \[\begin{array}{r l}
		    \textit{infoAdded}: \St &\longrightarrow \St\\
		    \sigma &\longrightarrow \sigma[\textit{infoSent} \mapsto \textit{infoSent} \+ 1] \\[.5em]
            \textit{resetLoop}: \St &\longrightarrow \St\\
		    \sigma &\longrightarrow \sigma[\textit{paths} \mapsto [~]][\textit{edges} \mapsto [~]][\textit{costs} \mapsto [~]][\textit{infoSent} \mapsto 0]\\[.5em]
		    \textit{addPath}_k: \St &\longrightarrow \St \hspace*{2em} \forall k \in \{1,\cdots,m\}\\
		    \sigma &\longrightarrow \sigma[\textit{paths} \mapsto \textit{paths}'][\textit{edges} \mapsto \textit{edges}'][\textit{costs} \mapsto \textit{costs}'] \\
		    \multicolumn{2}{c}{\textit{paths'} = \textit{paths} +\!+\textit{path}_k,\ \textit{edges}' = \textit{edges} +\!+ \textit{edges}_k,\ \textit{costs}' = \textit{costs} +\!+ L_k} \\[.5em]
		    \textit{reset}_k: \St &\longrightarrow \St\\
		    \sigma &\longrightarrow \sigma[\textit{path}_k \mapsto \textit{last}(\textit{path}_k)][\textit{edges}_k \mapsto [~]][L_k \mapsto 0]\\
		    \multicolumn{2}{c}{r(\tau_{i,j}) = (1-\rho)\cdot \tau_{i,j} + \Sigma_{k = 1}^m\Delta \tau_{i,j}^k \textrm{ with }
	    \Delta \tau_{i,j}^k = \left\{ \begin{array}{ll}
		                                \frac{Q}{\textit{costs}[k]}&   \textrm{if} \hspace{1.5mm} e_{i,j} \in \textit{edges}[k] \\
		                                0  &   \textrm{otherwise}\\
	                                  \end{array} \right.}
	   	\end{array} \]}	
	   	
		\hrulefill
	    \caption{Transformations of AS with m free ants.}
	    \label{tab:trans2}
	    \end{table}

\begin{table}[h]
	    \centering
	    \hrulefill
	    {

	    \[
     \begin{array}{l c l}
		    \textit{maxInfo}(\sigma) =\left\{ \begin{array}{ll}
			    \textit{True} &   \textrm{if } \sigma(\textit{infoSent}) < m \\
			    \textit{False} &    \textrm{otherwise}\\
                \end{array} \right .
                & \hspace*{6em} &

            \end{array}
                \]}

        $\forall \sigma \in \St$ and $\forall k \in \{1,\cdots,m\}$

        \hrulefill
	    \caption{Conditions of AS with $m$ free ants.}
	    \label{tab:cond2}
	    \end{table}

 Now that not all ants are going to participate in every iteration, the graph needs to control the information it sends. Suppose it did not. In that case, if it sent the information $m+1$ times in the last iteration of the algorithm then, when it received the $m$-th path generated in that iteration, it would evaluate the condition $\textit{end}$ to check that the algorithm is finished and start notifying the ants to stop their search. However, the ant that has received information but has been the last to send it would remain blocked until the graph is willing to receive this information, and we are not interested in the process that represents the graph to be receiving states after completing the algorithm, because useful information could be overwritten. For this reason, we introduce the variable $\textit{infoSent}\in \nat$, the transformation $\textit{infoAdded}$ and the condition $\textit{maxInfo}$. The idea is that $\textit{infoSent}$ carries the number of times the graph has sent its information. Thus, when it sends information to an ant, it will execute the transformation $\textit{infoAdded}$ to increment that counter and will be able to send information only if $\textit{infoSent}$ is less than $m$, which will be checked by the condition $\textit{maxInfo}$.

On the other hand, as now the same ant can generate several paths in the same iteration, the graph will need to save the information associated with these paths so that they are not eliminated when the ant is restarted. To do this, we introduce the variables $\textit{paths}\in \textit{List}(\real^+)$, $\textit{edges}\in \textit{List}(\real^+)$, and $\textit{costs}\in \textit{List}(\real^+)$ and the transformations $\textit{resetLoop}$ and $\textit{addPath}_k$ for each $k \in \{1,\cdots,m\}$. In this way, each time an ant $k$ finishes a lap, $\textit{path}_k$, $\textit{edges}_k$, and $L_k$ will be added to the lists $\textit{paths}$, $\textit{edges}$, and $\textit{costs}$ respectively with the transformation $\textit{addPath}_k$. On the other hand, $\textit{resetLoop}$ will be the transformation that restarts these variables at the beginning of each new iteration.
	
In addition, as the ants will not wait for each other, they will not restart all at once, but the graph will restart them when they finish a lap. Therefore we replace the transformation $\textit{reset}$ with the transformations $\textit{reset}_k$ for each $k \in \{1,\cdots,m\}$.
	
Finally, it is necessary to redefine the function $r$ that updates the number of pheromones of an edge in the transformation $\textit{trailUpd}$, since the organization of the information has changed. That is, now instead of having $m$ variables $\textit{path}_k$, $\textit{edges}_k$ and $L_k$, we have the lists $\textit{paths}$, $\textit{edges}$ and $\textit{costs}$. In Table~\ref{tab:trans2} the notation $\textit{edges}[k]$ and $\textit{costs}[k]$ indicates the position $k$ of the lists $\textit{edges}$ and $\textit{costs}$ respectively, and $Q$ is the constant given by the programmer. Note that now the traces are not updated according to the $m$ ants, but according to the $m$ paths that have been generated, which could well belong to the same ant or several, but not necessarily to all. Also, now the condition $\textit{completeIt}$ will be true if the length of the list $\textit{paths}$ is $m$. 	

		\begin{table}[h]
	    \centering
	    \hrulefill
	    {
	    \vspace*{0.5em}
	
		k-th ANT
		\[\begin{array}{l}
            A^k = rec\ X.(\textit{grA}_k?(x);([\textit{end}_A]\textit{stop} + [\neg \textit{end}_A]A_{\textit{info}}^k))\\
	        A_{\textit{info}}^k = recY.(\oplus_{i,j \in I} [p_{i,j}^k]\textit{move}_{i,j}^k;([\textit{complete}_k]A_C^k \+ [\neg \textit{complete}_k]Y))\\
	        A_C^k = \textit{grA}_k!;X
	        \end{array}\]
	    GRAPH
	    \[\begin{array}{l}
	    G = rec\ X.([\neg \textit{maxInfo}]G_E \+ G_R \+ [\textit{completeIt}]G_C)\\
	    G_E = \Sigma_{k = 1}^m \textit{grA}_k!;\textit{infoAdded};X \\
	    G_R = \Sigma_{k = 1}^m \textit{grA}_k?(x);\textit{addPath}_k;\textit{reset}_k;X\\
	    G_C = \textit{trailUpd};\textit{newIt};([\textit{end}]\textit{finish};G_F \+ [\neg \textit{end}]\textit{resetLoop};X)\\
	    G_F = recY.(\Sigma_{k = 1}^m [\neg \textit{notified}_k]\textit{grA}_k!;\textit{notify}_k;Y \+ [\textit{allNotif}]\textit{stop})\\
	    \end{array}\]		}
	    \hrulefill
	    \caption{Process algebra of AS with m free ants.}
	    \label{tab:PA2}
	    \end{table}

	The definition of the processes of the ants and the graph is given in Table~\ref{tab:PA2}. The behavior of the processes that represent the ants remains the same. It is the behavior of the process that represents the graph the one that is modified. This is a recursive process that can execute three options:
	\begin{enumerate}
		\item to run the process $[\neg \textit{maxInfo}]G_E$. That is, if the information has not been sent $m$ times yet and an ant requests the information from the graph, then it will send the current state and will return to the beginning of the recursive process.
		\item to run the process $G_R$. This will occur if an ant $k$ sends information to the graph. If this happens, then it must be because that ant has completed a lap, so the information related to this path is added to the lists $\textit{paths}$, $\textit{edges}$, and $\textit{costs}$ with the transformation $\textit{addPath}_k$. Then, the variables $\textit{path}_k$, $\textit{edges}_k$, and $L_k$ are restarted so that the ant begins a new path. Note that the ant will get this information as soon as it restarts its own recursive process and asks the graph for the information. Finally, the recursive process is restarted.
		\item to run the process $[\textit{completeIt}]G_C$. This process evaluates the condition $\textit{completeIt}$ to see if an iteration has been completed. In that case, the traces are updated with the transformation $\textit{trailUpd}$ and the counter $\textit{numIt}$ is increased with $\textit{newIt}$. The condition $\textit{end}$ is then evaluated to see if the algorithm has finished. If it has, then it executes the transformation $\textit{finish}$  and the process $G_F$, which is the same as in the previous version of the AS, to tell the ants that they should not continue the search. If not, then it restarts the recursive process.
	\end{enumerate}
	Note that if an iteration has been completed, only option 3 can take place. Indeed, if we have completed an iteration then it means that the length of the list $\textit{paths}$ is $m$, so the graph must have sent its information $m$ times and the condition $[\neg \textit{maxInfo}]$ is false. In addition, as we prevent sending the information more than $m$ times, if the list $\textit{paths}$ has length $m$ then it must be because all the ants that have received information have already sent it back, so no ant will be able to send information to the graph and option 2 will not take place either.
	
	With all this, assuming $M = \{1,\cdots,m\}$, we can model the AS with $m$ free ants as:
	$$\|_{k \in M} \ll A_k,\sigma_k \gg \| \ll G,\sigma_G \gg $$
	where $\forall k \in M$\\
    \[{
	    \begin{array}{clcl c clcl}
	        \sigma_k : &\Var &-\rightarrow & \Data & & \sigma_G: &\Var &-\rightarrow &\Data\\
		    &\textit{path}_k &\longrightarrow &[v_{0,k}] & & &\textit{numIt} &\longrightarrow &0\\
		    &\textit{edges}_k &\longrightarrow &[~] & & &\textit{final} &\longrightarrow &0\\
		    &L_k &\longrightarrow &0 & & &\textit{paths} &\longrightarrow &[~]\\
		    & & & & & &\textit{edges} &\longrightarrow &[~]\\
		    & & & & & &\textit{costs} &\longrightarrow &[~]\\
            & & & & & &\textit{infoSent} &\longrightarrow &0\\
		    & & & & & &\textit{notified} &\longrightarrow &[~]\\
		    & & & & & &\tau_{i,j} &\longrightarrow &\tau_0 ~\forall i,j \in \{1,\cdots,n\}, i\neq j
	    \end{array}}\]
\end{subsection}
	
\begin{subsection}{Formalization of fine-grained ${\cal M}{\cal M}$AS}
	\label{form-mmas}
	
		To model both the standard $\mathcal{MM}$AS and the free ants approach all the above can be used by simply modifying the function $r$ in the update of the traces to take into account the elitist strategy of the ant with the best path and the bounds of the amount of pheromone discussed in Section~\ref{subsec:maxminAS}.
\end{subsection}
	
\begin{subsection}{Formalization of fine-grained ACS with m ants}
		We now introduce the version of the ACS. The variables, transformations and conditions are given in Table~\ref{tab:setsVTC3}.

        \begin{table}[h]
	    \centering
	    \hrulefill
	
	    {
	    \begin{tabular}{l}
		$\Var = V_1 \cup V_2 \cup V_3$ where:\\
		\begin{tabular}{l}
			$V_1 = \{\textit{numIt, waiting, final,  notified}\}$ \\
			$V_2 = \{\textit{path}_k, \textit{edges}_k \ | \ 1 \leq k \leq m \}$ \\
			$V_3 = \{\tau_{i,j} \ | \ i,j \in I, i \neq j\}$\\
		\end{tabular}\\
		$\Trans = T_1 \cup T_2 \cup T_3$ where:\\
		\begin{tabular}{l}
			$ T_1 = \{ \textit{newIt, trailUpd,  reset, unblock}\}$\\
			$T_2 = \{\textit{\textbf{localUpd}}_k, \textit{block}_k, \textit{notify}_k \ | \ 1 \leq k \leq m\}$\\
			$T_3 = \{\textit{move}_{i,j}^k \ | \ 1 \leq k \leq m, i,j \in I, i \neq j\}$\\
		\end{tabular} \\
		$Cond =  C_1 \cup C_2  $ where:\\
		\begin{tabular}{l}
			$C_1 = \{\textit{end, end}_A\textit{, completeIt, allNotif}\}$\\
			$C_2 = \{\textit{complete}_k, \textit{wait}_k, \textit{notified}_k \ | \ 1 \leq k \leq m\}$\\
	   \end{tabular}
		\end{tabular}}
		
		\hrulefill
	    \caption{Sets of variables, transformations, and conditions for ACS with m ants.}
	    \label{tab:setsVTC3}
	    \end{table}
		In the case of ACS, it is necessary to modify the behavior of the algorithm so that local updates are made. Thus, the process that represents each ant will move through the graph and check if the algorithm has finished. The graph will be responsible for carrying out the local updates and for analyzing if the ant has taken a complete lap, in which case the $\offline$ update will be executed with the transformation $\textit{trailUpd}$ modified according to formula~(\ref{rastros-acs}). In addition, the probabilities will be calculated as in $\mathcal{MM}$AS.

        \begin{table}[h]
	    \centering
	    \hrulefill
	    {
	    \[\begin{array}{r l}
			\textit{localUpd}_k: \St &\longrightarrow \St \ \ \forall k \in \{1,\cdots,m\} \\
			\sigma &\longmapsto \sigma[\tau_{i^*,j^*}\mapsto (1 - \varphi)\cdot \tau_{i^*,j^*} + \varphi \cdot \tau_0]
	   	\end{array} \]}	
        where $e_{i^*,j^*} = \textit{last}(\textit{edges}_k)$\\
		\hrulefill
	    \caption{Transformations of ACS with m ants.}
	    \label{tab:trans3}
	    \end{table}

		Thus, we add the set of transformations $\textit{localUpd}_k$ that we will define according to the formula (\ref{rastros-acs}) of Section~\ref{aco-acs}, that is, requiring $e_{i,j} = \textit{last}(\textit{edges}_k)$. That is, only the number of pheromones present in the last edge of $\textit{edges}_k$ of the ant $k$ is updated.

		The rest of the variables, transformations and conditions are the same as in the version of the AS with $m$ ants of the Section~\ref{form-as-m}. We  model the behavior of the ants and the graph in this version of the ACS in Table~\ref{tab:PA3}.
		
		\begin{table}[h]
	    \centering
	    \hrulefill
	    {
	    \vspace*{0.5em}

		k-th ANT
	    \[\begin{array}{l}		
		A^k = rec\ X.(\textit{grA}_k?(x);(~ [\textit{end}_A]\textit{stop} \+ [\neg \textit{end}_A]A_{\textit{info}}^k~)~)\\
		A_{\textit{info}}^k = \oplus_{i,j \in I}[p_{i,j}^k]\textit{move}_{i,j}^k; \textit{grA}_k!;X\\
		 \end{array}\]
		
		GRAPH
	    \[\begin{array}{l}
	    G = rec\ X.(~G_E \+ G_R \+ [\textit{completeIt}]G_C~)\\
		G_E = \Sigma_{k = 1}^n [\neg \textit{wait}_k] \textit{grA}_k!;X \\
		G_R = \Sigma_{k = 1}^m \textit{grA}_k?(x);\textit{localUpd}_k; ( [\textit{complete}_k]\textit{block}_k;X \+ [\neg \textit{complete}_k];X ~)\\
		G_C = \textit{trailUpd}; \textit{newIt}; \textit{unblock}; (~[\textit{end}]G_F \+ [\neg \textit{end}]\textit{reset};X~)\\
		G_F = recY.(~\Sigma_{k = 1}^m [\neg \textit{notified}_k]\textit{grA}_k!;\textit{notify}_k;Y \+ [\textit{allNotif}]\textit{stop} ~)\\
	    \end{array}\]		}
	    \hrulefill
	    \caption{Process algebra for ACS with m ants.}
	    \label{tab:PA3}
	    \end{table}
		In this case, the process representing the ants is quite simple. After receiving the information of the graph, if the ant has finished then it stops its execution, whereas if the iteration has not been completed, then it is dedicated to moving through the graph probabilistically.

        The graph process is a recursive process that can perform three options:
		\begin{enumerate}
			\item to run the process $G_E$, which sends information to an ant that requests it and is not blocked,
			\item to run the process $G_R$, which receives information from an ant. In this case, it executes the corresponding $\textit{localUpd}_k$ transformation and evaluates the condition $\textit{complete}_k$. If the ant has completed a lap, then the process blocks it with the transformation $\textit{block}_k$ and starts again. Otherwise it simply returns to the beginning of the recursive process, letting the ant continue building its path.
			\item to run the process $[\textit{completeIt}]G_C$, which begins by evaluating the condition $\textit{completeIt}$ to know if the $\offline$ update of the traces must be carried out. Thus, if this condition is true then the process executes the following transformations: $\textit{trailUpd}$ to perform the $\offline$ update; $\textit{newIt}$ to increase the iteration counter; and $\textit{unblock}$, so that the ants are again available to receive information. Then, it evaluates the condition $end$ to see if the algorithm has finished. If not, then the variables $\textit{path}_k$, $\textit{edges}_k$, and $L_k$ are restarted for all $k \in \{1,\cdots,m\}$ and the process returns to the beginning. Note that it is necessary for the graph to perform this restart and to do so at this time. If it were done before evaluating the condition $\textit{end}$, then it could be true without all the ants having traveled the same path, so it would be enough that they had all ended up at the same vertex. In addition, if we restarted these variables and the algorithm ended, then we would not have preserved the paths of the ants and their lengths. On the other hand, if the algorithm has concluded, then the recursive process $G_F$ is executed to notify all the ants and finishes.
		\end{enumerate}
		Note that if $\textit{completeIt}$ is true then the process will not be able to run either option 1 or option 2, as with the AS. \\
		
        So, considering $M = \{1,\cdots,m\}$, the ACS algorithm can be formalized as follows:
		$$\|_{k \in M} \ll A_k,\sigma_k \gg \| \ll G,\sigma_G \gg $$
		where $\sigma_G$ and $\sigma_k$ are equal to the ones defined in Section~\ref{form-as-m}.
\end{subsection}

\begin{subsection}{Formalization of fine-grained ACS with $m$ free ants}\label{sec:ACSfree}

As in the case of the transition from the original AS to the AS with free ants, in this version of the ACS we also need to control the information that is sent. However, now we are not going to send the information $m$ times in each iteration, but we will send it $m\cdot n$ times, that is, each time an ant has to choose a new edge.

    \begin{table}[h]
	    \centering
	    \hrulefill
	
	    {
	    \begin{tabular}{l}
            $\Var = V_1 \cup V_2 \cup V_3$ where: \\
            \begin{tabular}{l}
	            $V_1 = \{\textit{numIt}, \textit{final}, \textit{paths}, \textit{edges}, \textit{costs}, \textit{notified},  \textit{\textbf{search}}, \textit{infoSent}\}$ \\
	            $V_2 = \{\textit{path}_k, \textit{edges}_k, L_k \ | \ 1 \leq k \leq m \}$\\
	            $V_3 = \{\tau_{i,j} \ | \ i,j \in I, i \neq j\}$\\
            \end{tabular}\\
            $\Trans =  T_ 1 \cup T_2 \cup T_3$ where: \\
            \begin{tabular}{l}
	            $T_1 = \{\textit{trailUpd}, \textit{newIt}, \textit{resetLoop}, \textit{infoAdded}\}$ \\
	            $T_2 = \{\textit{localUpd}_k, \textit{reset}_k, \textit{notify}_k, \textit{\textbf{addSearch}}_k, \textit{\textbf{eraseSearch}}_k\ | \ 1 \leq k \leq m\}$ \\
	            $T_3 = \{\textit{move}_{i,j}^k \ | \ 1 \leq k \leq m, i,j \in I, i \neq j\}$\\
            \end{tabular} \\
            $\textit{Cond} = C_1 \cup C_2 $ where: \\
            \begin{tabular}{l}
	            $C_1 = \{\textit{end}, \textit{end}_A, \textit{completeIt}, \textit{allNotif}, \textit{maxInfo}\}$ \\
	            $C_2 = \{\textit{complete}_k, \textit{notified}_k, \textit{\textbf{searching}}_k \ | \ 1 \leq k \leq m\}$ \\
	        \end{tabular}
		\end{tabular}}
		
	    \hrulefill
	    \caption{Sets of variables, transformations, and conditions of ACS with m free ants.}
	    \label{tab:setsVTC4}
	\end{table}
	
	In Table~\ref{tab:setsVTC4} the sets of variables, transformations, and conditions are given for this case. The change to take into account cannot be simply to consider that the information should be sent only $m\cdot n$ times, since this would not guarantee that it would always be sent to the same $m$ ants. By 'ants' here we mean iterations of some ant. Remember that when ants do not wait for each other, the $m$ paths do not have to be built by $m$ different ants, but could even come from a single ant. To solve this problem, we introduce the variable $\textit{search}$, the transformation sets $\textit{addSearch}_k$ and $\textit{eraseSearch}_k$, and the conditions $\textit{searching}_k$ for each $k \in \{1,\cdots,m\}$. Thus, $\textit{search}$ will carry the list of ants that have already begun to travel a lap, and $\textit{searching}_k$ will tell us if the k-th ant is on the list $\textit{search}$. These new transformations  and condition  are defined in Table~\ref{tab:trans4}  and Table~\ref{tab:cond3} respectively.

	    \begin{table}[h]
	    \centering
	    \hrulefill
	    {
	    \[\begin{array}{r l}
        	\textit{addSearch}_k: \St &\longrightarrow \St\\
	        \sigma &\longrightarrow  \sigma[\textit{search}\mapsto \textit{search} +\!+ k] \\
            \textit{eraseSearch}_k: \St &\longrightarrow \St\\
	        \sigma &\longrightarrow  \sigma[\textit{search} \mapsto \textit{remove}(k, \textit{search})] \\
	        \textit{resetLoop}: \St &\longrightarrow  \St\\
	        \sigma &\longrightarrow  \sigma' \\
	        \multicolumn{2}{c}{\sigma' = \sigma[\textit{paths} \mapsto [~]][\textit{edges} \mapsto [~]][\textit{costs} \mapsto [~]][\textit{infoSent} \mapsto 0][\textit{search }\mapsto [~]]}
	   	\end{array} \]}	
		\hrulefill
	    \caption{Transformations of ACS with m free ants.}
	    \label{tab:trans4}
	    \end{table}

 \begin{table}[h]
	    \centering
	    \hrulefill
	    {

	    \[
     \begin{array}{l c l}
		    \textit{searching}_k(\sigma) =\left\{ \begin{array}{ll}
			    \textit{True} &   \textrm{if } k \in \textit{search} \\
			    \textit{False} &    \textrm{otherwise}\\
                \end{array} \right .
                & \hspace*{6em} &

            \end{array}
                \]}

        $\forall \sigma \in \St$ and $\forall k \in \{1,\cdots,m\}$

        \hrulefill
	    \caption{Conditions of ACS with $m$ free ants.}
	    \label{tab:cond3}
	    \end{table}

In this way, when an ant begins to search, we will add it to the list $\textit{search}$ with the transformation $\textit{addSearch}_k$ and increase by one the counter $\textit{infoSent}$ with the transformation $\textit{infoAdded}$. When an ant completes a lap, we will remove it from the list with the transformation $\textit{eraseSearch}_k$, where the function $\textit{remove(elem,l)}$ removes the item $\textit{elem}$ from the list $l$. Thus, $\textit{search}$ will tell us who is going through the graph at all times, and the variable $\textit{infoSent}$ will tell us how many ants (or iterations of some ant) have done a search in this iteration of the algorithm.
In addition, in this case the transformation $\textit{resetLoop}$ will also have to restart the list $\textit{search}$.

    \begin{table}[h]
	    \centering
	    \hrulefill
	    {\footnotesize	
	    \vspace*{0.5em}

		k-th ANT
	    \[\begin{array}{l}		
            A^k = rec\ X.(~\textit{grA}_k?(x);(~ [\textit{end}_A]\textit{stop} \+ [\neg \textit{end}_A]A_{\textit{info}}^k~)~)\\
            A_{\textit{info}}^k = \oplus_{i,j \in I}[p_{i,j}^k]\textit{move}_{i,j}^k; \textit{grA}_k!;X \\
        \end{array}\]

        GRAPH
        \[\begin{array}{l}		
                G = rec\ X.(G_E \+ G_R \+ [\textit{completeIt}]G_C)\\
                G_E = \Sigma_{k = 1}^m ([\textit{searching}_k]\textit{grA}_k!;X + [\neg \textit{searching}_k]G_{EN})\\
                G_{EN} = [\textit{maxInfo}]X+[\neg \textit{maxInfo}]\textit{grA}_k!;\textit{addSearch}_k;\textit{infoAdded};X\\
                G_R = \Sigma_{k = 1}^m \textit{grA}_k?(x);\textit{localUpd}_k([\textit{complete}_k]\textit{reset}_k;\textit{eraseSearch}_k;X+[\neg \textit{complete}_k]X)\\
                G_C = \textit{trailUpd}; \textit{newIt};([\textit{end}]G_F \+ [\neg \textit{end}]\textit{resetLoop};X)\\
                G_F = recY.(\Sigma_{k = 1}^m [\neg \textit{notified}_k]\textit{grA}_k!;\textit{notify}_k;Y + [\textit{allNotif}]\textit{stop})\\
	       \end{array}\]		}
	    \hrulefill
	    \caption{Process algebra for ACS with m free ants.}
	    \label{tab:PA4}
	\end{table}
	
In Table~\ref{tab:PA4} the processes for ants and graphs are defined. The processes of the ants remain the same as in the previous version, as they are dedicated to selecting edges and immediately sending the information to the graph until the algorithm ends.

On the other hand, the process that represents the graph is a recursive process that can execute three options:
\begin{enumerate}
	\item to run the process $G_E$, which is the process in charge of sending the information to the ants. If the k-th ant wants information from the graph, then the process $G_E$ evaluates the condition $\textit{searching}$ to know if this ant is already moving through the graph. If so, then the graph sends it its state and restarts the initial recursive process. If, on the other hand, the ant intends to start its lap, then the process $G_{EN}$ is executed. This process evaluates the condition $\textit{maxInfo}$ to know if the $m$ ants (or iteration of some ant) that will make the paths in this iteration have already been selected, in which case the recursive process is restarted. If, on the other hand, information can still be sent to new ants, then the information is sent to the ant, it is added to the list $\textit{search}$ with the transformation $\textit{addSearch}$, the counter $\textit{infoSent}$ is increased, and the recursive process is restarted.
	\item to run the process $G_R$, which receives information from an ant $k$. In this case, it executes the corresponding $\textit{localUpd}_k$ transformation and evaluates the condition $\textit{complete}_k$. If the ant has completed a lap, then the process restarts its variables with the corresponding $\textit{reset}_k$ transformation, removes the ant from the list $\textit{search}$ with the transformation $\textit{eraseSearch}_k$ and starts again. Otherwise, it simply returns to the beginning of the recursive process, letting the ant continue building its path.
	\item to run the process $[\textit{completeIt}]G_C$, that starts by evaluating the condition $\textit{completeIt}$ to find out if it is needed to perform the $\offline$ upgrade of the traces. Thus, if this condition is fulfilled, then the process executes the transformations $\textit{trailUpd}$ to perform the $\offline$ update and $\textit{newIt}$ to increase the counter of iterations. After that, the condition $\textit{end}$ is checked to know if the algorithm has finished. Otherwise, the transformation $\textit{resetLoop}$ is executed to restart the iteration variables and to start with the recursive process again. On the other hand, if the algorithm has finished, then the recursive process $G_F$ is executed to notify all the ants and it finishes.
\end{enumerate}
Note that if an iteration has been completed, then the process will not be able to execute either option 1 or option 2. Indeed, if a round has been completed, then it means that all the ants that were searching have finished their round, so after evaluating $\textit{complete}_k$ the $G_R$ process has removed them from the $\textit{search}$ list with the $\textit{addSearch}$ transformation. Option 2 would also not be feasible because, if all the ants have finished, then they have no information to send.\\

With all of this, considering $M = \{1,\cdots,m\}$ we can model the ACS with $m$ free ants as:
$$\|_{k \in M} \ll H_k,\sigma_k \gg \| \ll G,\sigma_G \gg $$
where $\forall k \in M$ we have that $\sigma_k$ is defined as in the previous case and $\sigma_G$ is defined as before but including the variable $\textit{search}$ that takes the value of the empty list. That is, $\sigma_k$ and $\sigma_G$ are the states with the initialized values of the variables.


\end{subsection}


\section{Formalization of the coarse-grained parallelization of ACO }\label{coarse}\label{sec:coarse-grained}

A \textit{coarse-grained} parallelization consists in assigning large subpopulations to different processors with low communication among them.
As we saw in Section~\ref{subsec:parallelACO}, it has been empirically proven in \cite{manfrin06} that the way of implementing ACO that provides best results is the independent execution of different copies of the same sequential algorithm $\mathcal{MM}$AS with different initializations of the variables. In this implementation, a maximum number of iterations is set and the copies of the algorithm are executed without any communication between them. When they all have finished, the best solution among those provided by the different copies is chosen.
As this has been shown to be the most efficient strategy, we will formalize this parallelization approach for AS, ${\cal M}{\cal M}$AS and ACS. Additionally, in order to illustrate how to formalize any implementation in the spectrum from the most radical \textit{fine-grained} approach to the most radical \textit{coarse-grained} one, at the end of this section we will also discuss how to formalize \textit{coarse-grained} parallelizations with occasional communication.

\begin{subsection}{Formalization of coarse-grained AS with $p$ independent executions}\label{form-as-ci}
The idea of this section is to specify each copy of the sequential AS algorithm with a process $C$, using the maximum number of iterations as the stopping criterion. Once all the copies have finished, they will send their information to the process $B$, that will receive their solutions and select the best of them. Note that, as now we are specifying the sequential algorithm, there is no need to make any concrete distinction between the ants and the graph as we did in the previous sections.

Let $n$ be the number of vertices in the TSP, $p+1$ the number of processors available to solve the problem, and  $m$ the number of ants considered for the algorithm. Let $I = \{1,\cdots,n\}$, $S = \{1,\cdots,p\}$,  and $M = \{1,\cdots,m\}$. Each copy of the algorithm $C_s$ will be associated to the processor $s\in S$ and $B$ will be associated to the processor $p+1$.

The variables, transformations and conditions used for the formalization of the copies $C_s$ for all $s\in S$ and $B$ are given in Table~\ref{tab:setsVTC61}.

    \begin{table}[h]
	    \centering
	    \hrulefill

	    {
	    \begin{tabular}{l}
            $\Data = \mathbb{R^+} \cup List(\mathbb{R^+})$\\
            $\Var = V_1 \cup V_2 \cup V_3$ where:  \\
            \begin{tabular}{l}
	            $V_1 = \{\textit{numIt},
                \textbf{\textit{solutions}}, \textbf{\textit{result}}\}$ \\
	            $V_2 = \{\textit{path}_k, \textit{edges}_k,L_k \ | \ k \in M\}$ \\
	            $V_3 = \{\tau_{i,j} \ | \ i,j \in \{1,\cdots,n\}, i \neq j\}$ \\
            \end{tabular} \\
            $\Trans = T_1 \cup T_2$ where: \\
            \begin{tabular}{l}
	            $T_1 = \{\textit{trailUpd}, reset, \textit{newIt},
                \textbf{\textit{calculateBest}}\} \cup \{\textbf{\textit{selBest}}_s, \textbf{\textit{addSol}}_s | s\in S\}$ \\
    	        $T_2 = \{\textit{move}_{i,j}^k \ | \ k \in M, i,j \in I, i \neq j\}$\\
            \end{tabular}\\
            $Cond = \{\textbf{\textit{complete}}, \textit{end}, \textbf{\textit{allSolutions}}\}$\\
         \end{tabular}}

	    \hrulefill
	    \caption{Sets of variables, transformations, and conditions for \textit{coarse-grained} parallelization of AS with $p$ independent executions.}
	    \label{tab:setsVTC61}
	\end{table}

The variables $solution_s \in \textit{List}(\real^+)$ will keep the solution of the copies $C_s$ for all $s \in S$, and process $B$ will keep the solutions it receives from the other processes in $\textit{solutions} \in \textit{List(List}(\real^+))$ and store the best one in $\textit{result} \in \textit{List}(\real^+)$.

With regard to transformations, $\textit{trailUpd}$ will follow the elitist strategy of the ant with the best path and the bounded amount of pheromone discussed in  Section~\ref{subsec:maxminAS}. Also, we add the transformations $\textit{selBest}_s$ $\forall s \in S$ so that each copy of the algorithm chooses the best path among those provided by all the ants once it has finished. This new transformation is defined in Table~\ref{tab:trans5}, where $k^* \in M$ is the index of the ant that has built a better path, that is, $k^* = \textit{argmin}_{k \in M}\{L_k\}
$. Note that the variable $\textit{solution}_s$ is assigned a list whose first element is the cost of the best path, and the rest of the list is precisely the best path. This will help us compare the solutions of all the independently executed algorithms.

\begin{table}[h]
    \hrulefill
        {\footnotesize
    \[
    \begin{array}{r l}
            \textit{selBest}_s: \St &\longrightarrow \St\\
            \sigma &\longrightarrow \sigma[\textit{solution}_s \mapsto [L_{k^*}]+\!+\textit{path}_{k^*}]
            \hspace*{2em} \textrm{ where } k^* = \textit{argmin}_{k \in M}\{L_k\}\\
            	\textit{addSol}_s: \St &\longrightarrow \St\\
        \sigma &\longrightarrow \sigma[\textit{solutions} \mapsto \textit{solutions} +\!+ \textit{solution}_s ]\\
        \textit{calculateBest}: \St &\longrightarrow \St\\
        \sigma &\longrightarrow \sigma[\textit{result} \mapsto \textit{solutions}[k^*]]
        \hspace*{2em} \textrm{ where } k^* = \textit{argmin}_{k \in M}\{L_k\}
    \end{array}
    \]}	
    \hrulefill
    \caption{Transformations for $C_s$ and $B$.}
    \label{tab:trans5}    \label{tab:trans6}

\end{table}

We include the transformations $\textit{addSol}_s$, for all $s \in S$, and $\textit{calculateBest}$ so that $B$ selects the best solution. This way, when it receives the solution of the copy $s$ through the channel $\textit{sols}_s$, it will add the list $\textit{solution}_s$ to the list $\textit{solutions}$ with the transformation $\textit{addSol}_s$. When all the solutions have been received, it will select the best one with the transformation $\textit{calculateBest}$ and will assign it to the variable $\textit{result}$. The definitions of $\textit{addSol}_s$, $\forall s\in S$, and $\textit{calculateBest}$ are also given in Table~\ref{tab:trans6}, where $k^* \in M$ is the index of the list $\textit{solutions}$ corresponding to the best solution. That is, $k^* = \textit{argmin}_{k\in M} \{\textit{solutions}[k][1]\}$, where the index of the first element of a list is considered to be~1.
The condition $\textit{allSolutions}$ will be true when $B$ has received all the solutions, that is, the length of the list $\textit{solutions}$ is $m$ and is formally defined in Table~\ref{tab:cond4}. Finally, the condition \textit{complete} will tell each copy of the algorithm if an iteration is completed. Note that, as this is a sequential algorithm, the condition will be true when the length of any of the paths of the ants is $n$ (e.g. the ant 1).\label{cambio3}

\begin{table}[h]
 \centering
   \hrulefill
    {
    \[
      \begin{array}{l c l}
      \textit{allSolutions}(\sigma) =\left\{ \begin{array}{ll}
	               \textit{True} &   \textrm{if } \textit{len}(\sigma(\textit{solutions})) = m \\
	               \textit{False} &    \textrm{otherwise}\\
                    \end{array} \right .
            & \hspace*{6em} & \\
      \textit{complete}(\sigma) =\left\{ \begin{array}{ll}
			    \textit{True} &   \textrm{if } \textit{len}(\sigma(\textit{path}_1)) = n \\
			    \textit{False} &    \textrm{otherwise}\\
                \end{array} \right .
                & \hspace*{6em} &

      \end{array}
  \]}

  $\forall \sigma \in \St$

        \hrulefill
	\caption{Conditions of \textit{coarse-grained} parallelization             of AS with p independent executions.}
	\label{tab:cond4}

\end{table}

With these sets, we formalize the processes $C_s$, $\forall s \in S$, and $B$ as in Table~\ref{tab:PA5}. Thus, $C_s$ is a recursive process that starts by moving every ant to an edge in a probabilistic way. It then evaluates the $\textit{complete}$ condition to see if an iteration has been completed. If not, then the recursive process is restarted to choose new edges. If an iteration has been completed, then the traces are updated, the $\textit{numIt}$ counter is incremented, and the $\textit{end}$ condition is evaluated to see if the algorithm has finished, in which case the best solution is selected, sent through the channel $\textit{sols}_s$ and the process ends. If, on the other hand, the algorithm has not finished, then the recursive process is restarted for another iteration.

\begin{table}[h]
    \centering
    \hrulefill

    {\footnotesize	
	    \[\begin{array}{l}		
 \multicolumn{1}{c}{\textrm{s-th COPY OF THE ALGORITHM}} \\[0.5em]
           C_s = rec\ X.(\oplus_{i,j \in I} [p_{i,j}^1]\textit{move}_{i,j}^{1};\cdots;\oplus_{i,j \in I} [p_{i,j}^m]\textit{move}_{i,j}^{m};([\textit{complete}]C_C^s + [\neg \textit{complete}]X))\\
            C_C^s = \textit{trailUpd};\textit{newIt};([\textit{end}]\textit{selBest}_s;\textit{sols}_s!;\textit{stop} + [\neg \textit{end}]\textit{reset};X)\\[0.5em]
\multicolumn{1}{c}{\textrm{PROCESS B}} \\[0.5em]
            B = rec\ X.(~\Sigma_{s\in S}\textit{sols}_s?(x);\textit{addSol}_s;X \+ [\textit{allSolutions}]\textit{calculateBest};\textit{stop})\\
            \end{array} \]
            }
    \hrulefill
    \caption{\textit{Coarse-grained} formalization of AS with $p$ independent executions.}
    \label{tab:PA5}
\end{table}

The process $B$ receives the solutions of the others and adds them to the list $\textit{solutions}$, and then selects the best one to store it in the variable $\textit{result}$ and finishes.

Taking into account all of the above, we can model this implementation of the $\mathcal{MM}$AS in parallel as:
$$\|_{s \in S} \ll C_s, \sigma_s \gg \| \ll B,\sigma_B \gg $$
where for each $s \in S$ we have:
\[
    {
	\begin{array}{clcl c clcl}
	    \sigma_s : &\Var &-\rightarrow & \Data & & \sigma_B: &\Var &-\rightarrow &\Data\\
                &\textit{numIt} &\longrightarrow &0   & &  &\textit{solutions} &\longrightarrow &[~]\\
	        &\tau_{i,j} &\longrightarrow &\tau_0^s &\forall i,j \in I, i\neq j & &\textit{result} &\longrightarrow &[~]  \\
	        &\textit{path}_k &\longrightarrow &[v_{0,k}^s]~~ &\forall k \in M & &&&\\
	        &\textit{edges}_k &\longrightarrow &[~] &\forall k \in M & &&& \\
	        &L_k &\longrightarrow &0 &\forall k \in M & & &&\\
        \end{array}}
\]

\noindent Let us note that the values of the initialized variables are different (or not necessarily the same) in each copy of the algorithm.

\end{subsection}

\begin{subsection}{Formalization of coarse-grained ${\cal M}{\cal M}$AS with $p$ independent executions}\label{form-mmas-ci}
The ${\cal M}{\cal M}$AS implementation of these $p$ independent executions can be formalized almost as explained in the previous section. We just have to modify the function $r$ in the update of the traces to take into account the elitist strategy of the ant with the best path and the bounds of the amount of pheromone discussed in Section~\ref{subsec:maxminAS}.
\end{subsection}

\begin{subsection}{Formalization of coarse-grained ACS with $p$ independent executions}\label{form-acs-ci}

The formalization of the ACS in this scenario of $p$ independent copies of the algorithm is also very similar to the previous ones, as we only need to add the local updates and consider the appropriate function for the calculation of probabilities (see Section~\ref{subsec:ACS}). Thus, the variables, transformations and conditions used for this formalization are those in Table~\ref{tab:setsVTC61} together with the additional transformations $\textit{localUpd}_k$, $\forall k \in \{1,\cdots, m\}$, previously defined in Table~\ref{tab:trans3}.

\begin{table}[h]
    \centering
    \hrulefill

    {\tiny
     \[\begin{array}{l}	
\multicolumn{1}{c}{\textrm{s-th COPY OF THE ALGORITHM}} \\[0.5em]

            C_s = rec\ X.(\oplus_{i,j \in I} [p_{i,j}^1]\textit{move}_{i,j}^{1};\cdots;\oplus_{i,j \in I} [p_{i,j}^m]\textit{move}_{i,j}^{m};\textit{localUpd}_1;...;\textit{localUpd}_m;([\textit{complete}]C_C^s + [\neg \textit{complete}]X))\\
            C_C^s = \textit{trailUpd};\textit{newIt};([\textit{end}]\textit{selBest}_s;\textit{sols}_s!;\textit{stop} + [\neg \textit{end}]\textit{reset};X)\\[0.5em]

\multicolumn{1}{c}{\textrm{PROCESS B}} \\[0.5em]
B = rec\ X.(~\Sigma_{s\in S}\textit{sols}_s?(x);\textit{addSol}_s;X \+ [\textit{allSolutions}]\textit{calculateBest};\textit{stop})\\
            \end{array} \]
            }
    \hrulefill
    \caption{\textit{Coarse-grained} formalization of ACS with $p$ independent executions.}
    \label{tab:PA7}
\end{table}

In this case, the processes representing each copy $C_s$, $s \in S$, and the process $B$ in charge of selecting the final solution can be described as in Table~\ref{tab:PA7}, where the local updates introduced in the ACS are executed in each process $C_s$ after all the ants have selected the next vertex and $\textit{trailUpd}$ considers the function $r$ explained in Section~\ref{aco-acs}. The overall implementation of the ACS can be modelled as:

$$\|_{s \in S} \ll C_s, \sigma_s \gg \| \ll B,\sigma_B \gg $$
where for each $s \in S$ we have:
\[
    {
	\begin{array}{clcl c clcl}
	    \sigma_s : &\Var &-\rightarrow & \Data & & \sigma_B: &\Var &-\rightarrow &\Data\\
                &\textit{numIt} &\longrightarrow &0   & &  &\textit{solutions} &\longrightarrow &[~]\\
	        &\tau_{i,j} &\longrightarrow &\tau_0^s &\forall i,j \in I, i\neq j & &\textit{result} &\longrightarrow &[~]  \\
	        &\textit{path}_k &\longrightarrow &[v_{0,k}^s]~~ &\forall k \in M & &&&\\
	        &\textit{edges}_k &\longrightarrow &[~] &\forall k \in M & &&& \\
	        &L_k &\longrightarrow &0 &\forall k \in M & & &&\\
        \end{array}}
\]

\noindent Again, that the values of the initialized variables are different (or not necessarily the same) in each copy of the algorithm.
\end{subsection}

\begin{subsection}{Formalization of \textit{coarse-grained} parallelizations with occasional communication }\label{form-coarse-oc}
In the previous sections we have formalized the most "radical" \textit{coarse-grained} approach with no communication. The formalization given in this section aims to give an idea of how to formalize \textit{coarse-grained} parallelizations that consider occasional communication among their processes. A designer willing to construct a specification at same custom level of coarse granularity like this should provide the required details by using  the tools  given in this paper.

We will consider several copies of the sequential AS (the same idea would work with ${\cal M}{\cal M}$AS and ACS) and, this time, after a fixed number of iterations, the copies will share their traces (i.e. pheromone trails) and update them based on each others'. We will consider a general function $f^*$ for this update so that a designer can choose the concrete one that works better for the intended approach. Some typical examples of this sharing function are:
\begin{itemize}
    \item to update the traces with the traces of the copy with the best solution so far,
    \item to update the traces with the average of the copies, or
    \item to update the traces with a weighted average, typically prioritizing their own or the best.
\end{itemize}
Each copy will finish when the fixed maximum number of iterations \textit{maxIt} is reached.

The logic we will follow to formalize this implementation is a mix between the idea behind the \textit{fine-grained} cases and the previous \textit{coarse-grained} approach: there will be $p$ copies of the sequential algorithm as in the \textit{coarse-grained} one and a process $G^*$ behaving similarly to the graph of the \textit{fine-grained} one, now acting as a "super-graph" carrying the information of the graphs of the different copies. This process will be in charge of receiving the traces of each copy of the algorithm and will apply the corresponding $f^*$ function. It will also be responsible for the finalization of the rest of the processes.

Let $n$ be the number of vertices in the TSP, $p+1$ the number of processors available to solve the problem, $m$ the number of ants considered for the algorithm and $u$ the number of iterations each copy of the algorithm will execute before sharing their traces. Let $I = \{1,\cdots,n\}$, $S = \{1,\cdots,p\}$,  and $M = \{1,\cdots,m\}$. Each copy of the algorithm $C_s$ will be associated to the processor $s\in S$ and $G^*$ will be associated to the processor $p+1$.

The variables, transformations and conditions used for this formalization are given in Table~\ref{tab:setsVTC10}. Note that some of them have been previously defined in the \textit{fine-grained} implementations (see Section~\ref{sec:fine-grained}) but referring to the ants (e.g. $\textit{block}_k$) instead of to the copies. The necessary adaptations and changes are not difficult and the details will be omitted in this case.

\begin{table}[h]
		
	    \hrulefill
	
	    \centering
	    {
	    \begin{tabular}{l}
		$\Var = V_1 \cup V_2 \cup V_3$ where:\\
		\begin{tabular}{l}
			$V_1 = \{\textit{numIt, waiting, final, notified}\}$ \\
			$V_2 = \{\textit{path}_k^s, \textit{edges}_k^s, L_k^s | 1 \leq k \leq m, 1 \leq s \leq p\}$ \\
			$V_3 = \{\tau_{i,j}^s | 1 \leq s \leq p, i,j \in I, i \neq j\}$
		\end{tabular} \\
		$\Trans = T_1 \cup T_2 \cup T_3$ where:\\
		\begin{tabular}{l}
			$ T_1 = \{\textit{newIt, } \textbf{\textit{trailUpd}}_s \textit{, \textbf{trailUpd}}^*\textit{, reset, unblock, finish}\}$\\
			$T_2 = \{\textit{block}_s, \textit{notify}_s | 1 \leq s \leq p\}$\\
			$T_3 = \{\textit{move}_{i,j}^k | 1 \leq k \leq m, i,j \in I, i \neq j\}$\\
		 \end{tabular} \\
		$Cond =  C_1 \cup C_2  $ where:\\
		\begin{tabular}{l}
			$C_1 = \{\textit{end, end}_A\textit{, complete, completeIt, allNotif, \textbf{shareIt}}\}$ \\
			$C_2 = \{\textit{wait}_s, \textit{notified}_s | 1 \leq s \leq p\}$ \\
		\end{tabular}
		\end{tabular}}
		
		\hrulefill
	    \caption{Sets of variables, transformations and conditions for \textit{coarse-grained} parallelization of AS with occasional communication.}
	    \label{tab:setsVTC10}
	    \end{table}

In this implementation there are $p$ graphs (one for each copy of the algorithm), so each of them will have their variables for traces and will only update their own. That is why we introduce the transformations $\textit{trailUpd}_s$ $\forall s \in S$. Also, we include the transformation $\textit{trailUpd}^*$, that will apply the function $f^*$ to the traces. Both transformations are formally defined in Table \ref{tab:trans10}.

\begin{table}[h]
	    \centering
	    \hrulefill
	    {
	    \[\begin{array}{r l}
			\textit{trailUpd}_s: \St &\longrightarrow \St\\
			\sigma &\longrightarrow \sigma[\tau_{i,j}^s \mapsto r(\tau_{i,j}^s)] \\[0.5em]
			\multicolumn{2}{c}{r(\tau_{i,j}^s) = (1-\rho)\cdot \tau_{i,j}^s + \Sigma_{k = 1}^m\Delta             \tau_{i,j}^{s,k} \textrm{\ where \ }
		    \Delta \tau_{i,j}^{s,k} = \left\{ \begin{array}{ll}
			    \frac{Q}{L_k}&   \textrm{if } e_{i,j} \in \textit{edges}_k \\
			    0  &   \textrm{otherwise}\\

		    \end{array}
		    \right.} \\[0.5em]
            \textit{trailUpd}^*: \St &\longrightarrow \St\\
			\sigma &\longrightarrow \sigma[\tau_{i,j}^s \mapsto f^*(s,\tau_{i,j}^1,\cdots,\tau_{i,j}^p)] ~~\forall s \in S \\[0.5em]
	    \end{array} \]}
	    \hrulefill
	    \caption{Transformations of \textit{coarse-grained} formalization of AS with occasional communication.}
	    \label{tab:trans10}
	    \end{table}

Additionally, we add the condition $\textit{shareIt}$ that will tell each copy if it needs to send information to the super-graph, that is, if it made the next $u$ iterations or if it has reached the last iteration $\textit{maxIt}$ and the algorithm has finished. This condition is properly defined in Table \ref{tab:cond11}.
\begin{table}[h]
	    \centering
	    \hrulefill
	    {

	    \[
     \begin{array}{l c l}
		    \textit{shareIt}(\sigma) =\left\{ \begin{array}{ll}
			    \textit{True} &   \textrm{if } (\sigma(\textit{numIt}) = \textit{maxInfo}) \textrm{ or } (\sigma(\textit{numIt}) \textrm{ mod } \textit{u}  = 0) \\
			    \textit{False} &    \textrm{otherwise}\\
                \end{array} \right .
                & \hspace*{6em} &

            \end{array}
                \]}

        $\forall \sigma \in \St$ and $\forall k \in \{1,\cdots,m\}$

        \hrulefill
	    \caption{Conditions of \textit{coarse-grained} formalization of AS with occasional communication.}
	    \label{tab:cond11}
	    \end{table}

Based on all the above, the processes of the copies and the super-graph are defined in Table \ref{tab:PA10}.

		\begin{table}[h]
	    \centering
	    \hrulefill
	    {\footnotesize	
	    \vspace*{0.5em}
	
			s-th COPY OF THE ALGORITHM
		\[\begin{array}{l}
		     C_s= rec\ X.(\textit{channel}_s?(x);([\textit{end}_A]\textit{stop} + [\neg \textit{end}_A]C_{\textit{info}}^s))\\
		     C_{\textit{info}}^s = recY.(\oplus_{i,j \in I} [p_{i,j}^1]\textit{move}_{i,j}^{1};\cdots;\oplus_{i,j \in I} [p_{i,j}^m]\textit{move}_{i,j}^{m};([\textit{complete}]C_{send}^s + [\neg \textit{complete}]Y)\\
          C_{\textit{send}}^s = \textit{trailUpd}_s;\textit{newIt};\textit{reset};([\textit{shareIt}]\textit{channel}_s!;X \+ [\neg \textit{shareIt}]Y)
		\end{array} \]
	    	SUPER-GRAPH
		\[\begin{array}{l}
		    G^* = rec\ X.(\Sigma_{s = 1}^{p}[\neg \textit{wait}_s]\textit{channel}_s!;X  +  \Sigma_{s = 1}^p \textit{channel}_s?(x);\textit{block}_s;X + [\textit{completeIt}]G_C^*)\\
		    G_C^* = ([\textit{end}]\textit{finish};G_F \+ [\neg \textit{end}]\textit{trailUpd}^*;\textit{unblock};X)\\
		    G_F^* = recY.(\Sigma_{s = 1}^p [\neg \textit{notified}_s]\textit{channel}_s!;\textit{notify}_s;Y \+ [\textit{allNotif}]\textit{stop})\\
		    \end{array}\]
		}
	    \hrulefill
	    \caption{\textit{Coarse-grained} formalization of AS with occasional communication}
	    \label{tab:PA10}
	    \end{table}

Similarly to what we studied in Section~\ref{form-as-m} with the standard AS, the super-graph will send the traces of the different copies of the algorithm to each process $C_s$, $s \in S$, and each of the copies will execute $u$ iterations of the algorithm sequentially. Once they have finished this set of iterations, they will send their traces to the super-graph. Every time the super-graph receives the information of one copy, it will block it so that the rest of the copies can finish their $u$ iterations too.
When all the copies have finished this set of iterations, the super-graph applies the global updating function $f^*$ with the transformation $\textit{trailUpd}^*$, unblocks all the copies and sends them the new traces. The condition \textit{end} will tell the super-graph if the copies have finished the algorithm by comparing \textit{numIt} with the maximum number of iterations. Note that it will compare \textit{maxIt} with \textit{numIt}, that will have assigned the value of the last copy of the algorithm that sent its information. This is not a problem as all the copies wait for each other in each set of $u$ iterations, so they all send the same value of \textit{numIt} to the super-graph. When the condition \textit{end} is true, the super-graph will notify the copies they all have finished so that they stop their execution, just as we did in Section~\ref{form-as-m} with the standard AS.

Finally, note that we have omitted the part of the process where all the copies send their solutions to the super-graph so that it selects the best one as we did in the previous \textit{coarse-grained} parallelizations. It will be up to the designer to complete the formalization of this part according to the intended occasional communication approach, as it was explained previously.\label{cambio4}

The overall implementation of the AS can be modelled as:

$$\|_{s \in S} \ll C_s, \sigma_s \gg \| \ll G^*,\sigma_{G^*} \gg $$
where for each $s \in S$ we have:
\[
    {
	\begin{array}{clcl c clcl}
	    \sigma_s : &\Var &-\rightarrow & \Data & & \sigma_{G^*}: &\Var &-\rightarrow &\Data\\
                &\textit{numIt} &\longrightarrow &0   & &  &\textit{waiting} &\longrightarrow &[~]\\
	        &\tau_{i,j}^s &\longrightarrow &\tau_0^s &\forall i,j \in I, i\neq j & &\textit{notified} &\longrightarrow &[~]  \\
	        &\textit{path}_k^s &\longrightarrow &[v_{0,k}^s]~~ &\forall k \in M & & final&\longrightarrow&0\\
	        &\textit{edges}_k^s &\longrightarrow &[~] &\forall k \in M & &&& \\
	        &L_k^s &\longrightarrow &0 &\forall k \in M & & &&\\
        \end{array}}
\]

\noindent Again, the values of the initialized variables are generally different in each copy of the algorithm.

\end{subsection}


\section{Discussion: comparison of specifications}\label{sec:discussion}

In this section we summarize all the ACO variants presented in the paper, highlighting the specific aspects that make each case different to ease their mutual comparison.

In the {\it fine-grained} specifications of ACO, all ants are run in parallel along with the graph process. In this case we have considered the following specification variants:

\begin{itemize}
\item AS: in the fine-grained specification of AS, when an ant $k$ finishes its iteration, the graph blocks it with transformation $\textit{block}_k$. This transformation adds it to the $\textit{waiting}$ variable, which stores the ants which have already finished. The graph evaluates a condition to make sure it does not send information to any ant in this list, so that these ants cannot start their iteration. When all ants are in the $\textit{waiting}$ list, the graph unlocks them, restarts the variables, and the algorithm continues with the next iteration.

\item $\mathcal{MM}$AS: the fine-grained specification of $\mathcal{MM}$AS is like the previous one, but the elitist strategy and the corresponding trail update mechanism are introduced.

\item ACS: the fine-grained specification of ACS is similar to the corresponding one of AS, with the difference that the mechanism of the local update when the ant moves is introduced. In order to carry out these additional updates, the graph will receive the information of the ants every time each of them moves. Therefore, it will be responsible for counting when an iteration is completed to make the \textit{offline} updates.

\item AS with free ants: in this alternative version of the fine-grained specification of AS, faster ants are allowed to perform more paths per algorithm iteration. In order to make each iteration depend on the same number of new ant tours and guarantee that no ant performs a task that will not be taken into account, the graph process counts how much information it sends to ants, so that it sends a predefined number of requests of performing graph tours to all ants.

\item $\mathcal{MM}$AS with free ants: this is the $\mathcal{MM}$AS version of the previous variant.

\item ACS with free ants: this is the ACS version of the previous two variants. The peculiarity of this specification, with respect to the corresponding AS one, is that now we have local update operations, which implies sending data $n$ times to each ant in each iteration: every time an ant moves to another node, the graph sends the new pheromone trails to the ant.
Note that the strategy of just sending the data $n\cdot m$ times would not work, because it does not guarantee that the data would be sent to an ant which is already moving.
In order to properly cope with this problem, we add the variable $\textit{searching}$, which is a list where ants are included when the graph sends them the information for the first time in the current iteration.
This way, we know this particular ant has the {\it right} to receive information from the graph when it needs it, because it is actually performing a graph tour. The graph will eventually know when the ant has finished its iteration, and it will take the ant out of the $\textit{searching}$ list when the ant finishes.
Hence, if there is still work to be done in this iteration, this ant will be able to be selected again for a new round.
Every time the graph adds an ant to the $\textit{searching}$ list, the $\textit{maxInfo}$ counter will be increased.
Thus, when this counter reaches value $m$, no more ants can be added to this list (which grows and decreases as long as ants start and finish). This way, no ant performs unwanted work.
\end{itemize}

In the coarse-grained specifications of ACO, several copies of the sequential algorithm are used, and an additional process coordinates them. The following specification variants are considered in this case:

\begin{itemize}
\item AS without communication: in the coarse-grained specification of AS without communication, $p$ instances of the sequential algorithm are used. When these instances finish, each one sends its best result to an auxiliary process $B$, which just picks the best one of them.
\item $\mathcal{MM}$AS without communication: the coarse-grained variant of $\mathcal{MM}$AS without communication is similar to the previous one. Let us remind that, as stated in~\cite{manfrin06}, this is the best parallel strategy for ACO.
\item ACS without communication: this specification is very similar to the corresponding AS one. The only difference is that the local update transformation must be added. Since it is sequential, it must be introduced after the movement of the ants.
\item AS with occasional communication: in this variant of the coarse-grained specification of AS, again $p$ instances of the sequential algorithm are executed in parallel, although these instances share their pheromone trails once in some predefined number of iterations by means of a generic sharing function. This function can be defined in different ways: it may make all instances copy the pheromone trails of the algorithm instance having the best solution; it may make all pheromone trail of each edge become the average of pheromone trails of the same edge in all algorithm instances; it may make the pheromone trails be {\it partially} influenced by the pheromone trails of the best algorithm instance by using some weighted average; etc. This pheromone trails update operation is performed by
a process which plays the role of a ``super-graph''.
\end{itemize}

\section{Conclusions and future work}\label{sec:concl}

This paper has presented a process algebraic specification of several ACO algorithms with  two  parallelization granularities, constituting, to the best of our knowledge, the first full formal definition of their functional and concurrent behavior all together.
By constructing a process algebra with an appropriate structure for this goal, we could check the usefulness of process algebras to practically specify the algorithms in the domain under consideration.
In order to develop this language in such a way that our specifications in the language are natural, we faced several design choices such as e.g. the coordination between probabilistic and non-deterministic choices, how local information is kept and passed through processes, how this information affects probabilities, etc.
The resulting algebra was presented in Section~\ref{mialg}.
Next we showed how to use this algebra in the specification of processes within the chosen domain, in particular to specify the parallel ACO algorithms. We presented several types of implementation approaches, covering a wide range in the types of parallelization.

From this study, we consider several lines of future work. We wish to apply the developed algebra in the specification of other swarm intelligence algorithms. These algorithms are usually used in the resolution of complex and large problems, so it is increasingly common to implement them in parallel.
The proposed algebra allows us to naturally model different types of parallelization, which is suitable for these algorithms as well. Moreover, in many of them it is necessary to introduce probabilities in a very similar way as how we did in this work. We will study what changes would be necessary in our algebra, if any is actually needed, so that a high number of swarm intelligence algorithms can be covered without losing concreteness.
Algorithms such as RFD (\textit{River Formation Dynamics}) (\cite{rrrUC07}), PSO (\textit{Particle Swarm Optimization}) (\cite{poli07}), and ABC (\textit{Artificial Bee Colony}) (\cite{kumar09}) look particularly suitable to be specified with $PA^2CO$, both in their sequential form and under different types of parallelization.


\bigskip
\noindent\textbf{Credits:}
{\bf María García:} methodology, investigation, validation, formal analysis, writing - original draft. {\bf Natalia López:} conceptualization, supervision, project administration, writing - original draft and review \& editing. {\bf Ismael Rodríguez:} conceptualization, supervision, project administration, writing - original draft and review \& editing.


\bibliographystyle{elsarticle-harv}
\bibliography{bibliography}





\end{document}